\documentclass[english,conference,ruled]{IEEEtran}
\usepackage[T1]{fontenc}
\usepackage[latin9]{inputenc}
\usepackage{color}
\usepackage{babel}
\usepackage{algorithm2e}
\usepackage{amsmath}
\usepackage{amsthm}
\usepackage{amssymb}
\usepackage{graphicx}
\usepackage[unicode=true,pdfusetitle,
 bookmarks=true,bookmarksnumbered=false,bookmarksopen=false,
 breaklinks=false,pdfborder={0 0 1},backref=false,colorlinks=true]
 {hyperref}
\hypersetup{
 pdfborderstyle=}

\makeatletter
\theoremstyle{plain}
\newtheorem{thm}{\protect\theoremname}
\theoremstyle{remark}
\newtheorem{rem}[thm]{\protect\remarkname}

\usepackage{babel}
\newcommand{\nosemic}{\renewcommand{\@endalgocfline}{\relax}}
\newcommand{\dosemic}{\renewcommand{\@endalgocfline}{\algocf@endline}}
\usepackage{cite}
\usepackage[paper=letterpaper,top=0.75in,bottom=0.75in,right=0.75in,left=0.75in]{geometry}
\usepackage{subfigure}
\usepackage{enumerate}

\newtheorem{definition}{Definition}\newtheorem{theorem}{Theorem}

\providecommand{\remarkname}{Remark}
\providecommand{\theoremname}{Theorem}

\makeatletter



\makeatletter
\newcommand*\titleheader[1]{\gdef\@titleheader{#1}}
\AtBeginDocument{%
  \let\st@red@title\@title%
  \def\@title{%
    \bgroup\normalfont\large\centering\@titleheader\par\egroup
    \vskip1.5em\st@red@title}
}
\makeatother

\title{Stitching Dynamic Movement Primitives and Image-based Visual Servo Control}

\titleheader{This work has been submitted to the IEEE for possible publication. Copyright may be transferred without notice, after which this version may no longer be accessible.}

\begin{document}
\newgeometry{top=1in,bottom=0.75in,right=0.75in,left=0.75in}
\IEEEoverridecommandlockouts 
\author{Ghananeel Rotithor, \textit{Student Member, IEEE}, Iman Salehi, \textit{Student Member, IEEE}, \\Edward Tunstel, \textit{Fellow, IEEE},  Ashwin P. Dani, \textit{Senior Member, IEEE} \thanks{This work was supported in
part by a Space Technology Research Institutes grant (number 80NSSC19K1076)
from NASA Space Technology Research Grants Program and in part by NSF grant no. SMA-2134367. Ghananeel Rotithor, Iman Salehi, and Ashwin P. Dani are with the Department of Electrical
and Computer Engineering at University of Connecticut, Storrs, CT
06269. Email: \{ghananeel.rotithor; iman.salehi; ashwin.dani\}@uconn.edu. Edward Tunstel is with Raytheon Technologies Research Center, East Hartford, CT 06108. Email: tunstel@ieee.org.}}
\maketitle
\begin{abstract}
Utilizing perception for feedback control in combination with Dynamic Movement Primitive (DMP)-based motion generation for a robot's end-effector control is a useful solution for many robotic manufacturing tasks. For instance, while performing an insertion task when the hole or the recipient part is not visible in the eye-in-hand camera, a learning-based movement primitive method can be used to generate the end-effector path. Once the recipient part is in the field of view (FOV), Image-based Visual Servo (IBVS) can be used to control the motion of the robot. Inspired by such applications, this paper presents a generalized control scheme that switches between motion generation using DMP and IBVS control. To facilitate the design, a common state space representation for the DMP and the IBVS systems is first established. Stability analysis of the switched system using multiple Lyapunov functions shows that the state trajectories converge to a bound asymptotically. The developed method is validated by three real world experiments using the eye-in-hand configuration of a Baxter research robot.
\end{abstract}
\section{Introduction}

Robots are required to perform fine manipulation, navigation, and target tracking tasks in a variety of applications ranging from manufacturing automation to space exploration. To perform tasks involving fine manipulation, incorporating visual feedback can be beneficial when the robot end-effector is close to the object being manipulated. To perform tasks such as landing a quadcopter from a far away distance, incorporating visual feedback is useful when image features are visible to an onboard camera sensor. When the object is not in the field of view (FOV) of the camera or reliable features cannot be extracted or if there is a feature loss, learning-based motion generation methods such as dynamic movement primitives (DMP) can be used to control the robot. For example, in a wire pinning task for wire harness assembly, the robot holding the wire should reach the pin by following a certain trajectory that avoids collision with the fixtures holding the pin. The pin may not always be in the FOV of the camera attached to robot end-effector. In such cases, to insert the wire into the pin, a  learned DMP can be used to generate the motion that takes the wire near to the pin before the visual feedback of the pin is available \cite{Wirepinning}. Motivated by such examples and a need to ensure stable motion when switching between learned and vision-based controllers, a new methodology that unifies DMP \cite{ijspeert2003learning} and Image-based Visual Servoing (IBVS)\cite{chaumette2006visual} is proposed in this paper. The proposed algorithm enables stable switching between DMP and IBVS controllers to reach the goal location from an initial location when the visual feedback is not continuously available. The proposed switched system is referred to as a DMP-IBVS switched system. 

Many approaches have been developed to design robot controllers using visual feedback that keep the objects in the FOV (see, e.g., \cite{lopez2009homography,gans2011keeping,Salehi2021ICRA}). A trajectory planning and tracking control approach for mobile robots is developed in \cite{Wang2021SMC} that uses virtual-goal-guided rapidly exploring random tree (RRT) for trajectory planning and IBVS with the FOV constraints for tracking. A constrained IBVS approach for helicopter landing on a moving platform is proposed in \cite{Huang2021SMC} using a control barrier function methodology to satisfy the FOV constraints. A model predictive controller (MPC) for visual servoing of a mobile robot is developed in \cite{Ke2017SMC} which handles visibility constraints for IBVS and velocity limits of the robot. In contrast, a learning-based robot motion generation using DMP is developed in this paper to generate robot control commands in the absence of visual feedback, and a switching control law is developed to switch to IBVS controller when visual feedback is available. 

It is known that switching arbitrarily between stable subsystems can lead to instability\cite{liberzon2003switching,liberzon1999basic}. In \cite{hespanha1999stability}, an average dwell time condition based on multiple Lyapunov functions is developed for stabilizing switched systems. In \cite{daafouz2002stability}, the authors propose a linear matrix inequality-based condition to check for the existence of a quadratic Lyapunov function for proving the asymptotic stability of a switched discrete system. In \cite{hespanha2004uniform}, LaSalle's invariance principle is extended to switched linear systems to deduce asymptotic stability using multiple Lyapunov functions whose Lie derivatives are negative semidefinite. In \cite{lin2009stability}, a survey of results for the stability of switched linear systems is presented, and the problem of stabilizability of switched systems is analyzed. Multiple Lyapunov functions for analyzing Lyapunov stability and iterated function systems as a tool for Lagrange stability are proposed in \cite{branicky1998multiple}. Invariance-like results for nonautonomous nonlinear switched systems are developed in \cite{kamalapurkar2018invariance}. Switched systems analysis of output feedback systems and observers have been studied in the literature. The switching method in \cite{Astolfi2020_tac} analyzes switching between locally and globally converging observers with asymptotic stability. A composite output feedback control law is derived in \cite{Prieur2011_tac} that switches between locally and globally asymptotically stable output feedback controllers.  

Switched systems analysis has been used to solve challenging problems in robotics, including visual servo control and vision-based trajectory tracking. In \cite{gans2007stable} an asymptotically stable hybrid switched systems visual servo controller is proposed, which switches between IBVS and position-based visual servo control (PBVS) based on multiple Lyapunov functions. The switching between IBVS-PBVS control is developed to switch between unstable systems and increase the region of stability. An estimator that switches between state predictor and image-based observer for object localization is presented in \cite{parikh2016switched} when the image feedback is intermittently available. The stability of the switched system is established using a common Lyapunov function. An estimator for pose estimation of a moving target is developed in \cite{parikh2018target} when measurements are available intermittently by learning the motion model. In \cite{parikh2017switched,chen2019switched}, an observer-predictor framework is presented for target tracking and trajectory tracking in the presence of intermittent measurements by switching between a global uniformly ultimately bounded (GUUB) and an unstable system. The switched error system is shown to be GUUB based on the dwell time conditions. The approaches for image-based tracking use switching between systems mainly to accommodate feature track losses, occlusions, and limited camera field of view (FOV) of features. Switched systems analysis is also developed for target tracking in the presence of intermittent observations from a network of stationary cameras in \cite{harris2020target}. In \cite{zegers2019switched}, consensus protocols of a distributed multi-agent system are developed using switched systems analysis wherein the leader provides intermittent information to all the followers.
Application of switched system to biped locomotion is developed in \cite{veer2019switched} where the boundedness of input-to-state (ISS) stable switched systems with multiple equilibria is proven. The switched system results developed or applied in these results switch between systems that are either globally stable or locally stable.

In contrast to the switched systems approaches developed to handle feature track losses, occlusions in object tracking and localization, and approaches developed to improve the stability in IBVS-PBVS switching, this paper develops a new method that switches between image-based control and learning-based control when the image-feedback is not available such as in instances which may include limited camera FOV, occlusions, and feature losses. In this paper, a new switched systems analysis is developed for systems that switch between locally stable (IBVS system) \textit{and} globally stable (DMP system) modes. The goal is to regulate the camera pose with respect to a target. 
When visual feedback is available, the IBVS controller is used to generate camera accelerations. When visual feedback is not available, then DMP is used to generate camera accelerations. DMP is implemented as an online end-effector acceleration controller whose weights are trained using a pose regulation task demonstration data. The method can be useful in robotic tasks where a robot arm with eye-in-hand configuration is controlled using DMP and IBVS controllers for achieving precision or when a ground robot is controlled using a camera mounted on the robot with image and non-image-feedback. The technical contributions of this paper are as follows: 
\begin{itemize}
    \item A new common state-space representation is developed to analyze the switched system stability of the IBVS and DMP acceleration controllers.
    \item A new IBVS acceleration control law is developed and the corresponding closed-loop dynamics are proven to be UUB if the initial state is sufficiently close to the goal state.
    \item Combined position and orientation DMPs are presented and the corresponding dynamics are proven to be globally asymptotically stable.
    \item The switching between the locally stable IBVS controller and globally stable DMP is analyzed using multiple Lyapunov functions and the switched system dynamics are proven to converge asymptotically to a bound whose analytical expression is derived. Furthermore, an algorithm is developed based on the analysis of the switched system to ensure stable switching between the IBVS and DMP subsystems.
\end{itemize} 
Compared to our preliminary development in \cite{rotithor2020combining}, this paper provides rigorous stability analyses for the individual DMP and IBVS systems along with a method for switching and stability analysis for the switched system. Experimental results for a pose regulation task are presented using the eye-in-hand configuration of the Baxter robot, and the switched system results are validated in the presence of occlusions in the image-feedback, and compared with a DMP-only controller.

The rest of the paper is organized as follows. In Section \ref{sec:Coordinate-System}, the coordinate frames attached to the robot base, camera, the goal location and its kinematics are discussed. In Section \ref{sec:IBVS}, an acceleration controller is presented for IBVS and the error dynamics are proven to be uniformly ultimately bounded (UUB). In Section \ref{sec:TaskSpace-Control}, position and orientation DMPs are presented and global asymptotic stability of the error dynamics is presented. In Section \ref{sec:Hybrid-System}, the stability of the switched system is proven and an average dwell time condition is developed. Experimental results of the switched system are presented in Section \ref{sec:Experiments}.

\textit{Notations:} The set of real numbers and integers is denoted by $\mathbb{R}$ and $\mathbb{Z}$, respectively. The symbols $\mathbb{R}^{+}$ and $\mathbb{Z}^{+}$ denote the set of non-negative real numbers and non-negative integers, respectively. The standard Euclidean norm of a vector is denoted by $\Vert \cdot \Vert$ and for a $p$ dimensional real vector the open ball is defined as $\mathcal{B}_{\zeta}(x)=\left\{x' \in \mathbb{R}^{p}\: \big\vert\: \Vert x-x' \Vert < \zeta \right\}$, where $\zeta > 0$ is a constant. For a matrix $A\in \mathbb{R}^{n\times n}$, its symmetric part is denoted by $\mathrm{sym}\left\{A\right\}=\frac{1}{2}\left(A+A^{T}\right)$. 

\section{Problem Formulation\label{sec:Coordinate-System}}
\begin{figure}[h]
\begin{centering}
\includegraphics[width=0.8\columnwidth]{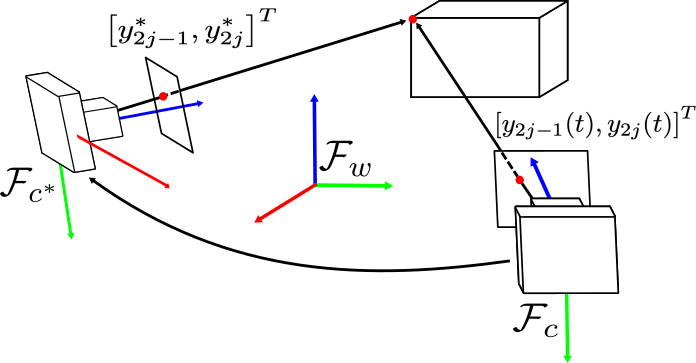}
\par\end{centering}
\begin{centering}
\caption{\label{fig:Reference-frames}Reference frames attached to the camera, the goal location and the inertial reference frame.}
\par\end{centering}
\end{figure}

Consider a fixed inertial reference frame $\mathcal{F}_{w}$ attached to the robot base and a coordinate frame $\mathcal{F}_{c}$ attached to a moving camera observing a static object shown in Fig. \ref{fig:Reference-frames}. The camera frame $\mathcal{F}_{c}$ is attached to the camera in such a way that the $Z$-axis of the coordinate frame aligns with the optical axis of the camera and the $X$ and $Y$ axes form the basis of the image plane. Thus, the coordinate frame attached to the camera always moves along with it, as the camera undergoes motion. Let $R_{b}^{a}\in SO(3)$ and $t_{b}^{a}\in\mathbb{R}^{3}$ be the rotation and translation from $a$ to $b$ expressed in $a$, respectively. A quaternion representation of the rotation matrix $R_{b}^{a}$ is given by $q_{_{b}}^{a}\in\mathcal{S}^{4}$, where $\mathcal{S}^{p}=\left\{ x\in\mathbb{R}^{p}\left|x^{T}x=1\right.\right\} $ is the unit hypersphere. It is assumed that the pose of the camera in the world reference frame, $T_{c}^{w}(t)$ can be measured. A coordinate frame $\mathcal{F}_{c^{*}}$ is attached to the desired goal location such that when $\mathcal{F}_{c}$ coincides with $\mathcal{F}_{c^{*}}$ the image feature error and the pose error is zero. Additionally, the pose transformation between $\mathcal{F}_{w}$ and $\mathcal{F}_{c^{*}}$, denoted as $T_{c^{*}}^{w}$, is assumed to be known. The time varying pose transformation between the goal camera frame and the current camera frame is denoted by $T_{c}^{c^{*}}(t)$ and is represented as $\left[(t_{c}^{c^{*}}(t))^{T},(q_{c}^{c^{*}}(t))^{T}\right]^{T}\in\mathbb{R}^{3}\times\mathcal{S}^{4}$.

For designing a switching control law between DMP and IBVS, a common state space is established between them. A new common auxiliary state is defined as follows
\begin{equation}
x(t)=\left[\begin{array}{cc}
e_{p}^{T}(t) & \xi_{c^{*}}^{T}(t)\end{array}\right]^{T}\in\mathbb{R}^{12},\label{eq:State}
\end{equation}
where $e_{p}(t)=\left[(t_{c}^{c^{*}}(t))^{T},r^{T}(t)\right]^{T}\in\mathbb{R}^{6}$ and $r(t)=2\mathrm{log}(q_{c}^{c^{*}}(t))=\varTheta(t)\mathbf{n}(t)$ is the quaternion logarithm that transforms a quaternion into the corresponding angle-axis product where $-\pi<\varTheta(t)<\pi$. The pose transformation $e_{p}(t)$ can be viewed as the pose error that indicates the deviation of the current camera frame $\mathcal{F}_{c}$ from the goal camera frame $\mathcal{F}_{c^*}$. In (\ref{eq:State}), $\xi_{c^{*}}(t)=\left[\begin{array}{cc}
v_{c^{*}}^{T} (t)& \omega_{c^{*}}^{T} (t)\end{array}\right]^{T}\in\mathbb{R}^{6}$ is the velocity of the camera expressed in the desired frame $\mathcal{F}_{c^{*}}$, such that $v_{c^{*}}(t)\in\mathbb{R}^{3}$ is the linear velocity and $\omega_{c^{*}}(t)\in\mathbb{R}^{3}$ is the angular velocity. \\
\textbf{Problem Description and Solution Approach:} Given the desired pose $T_{c^{*}}^{w}$, and the current camera pose $T_{c}^{w}(t)$, the problem is to regulate the current camera frame $\mathcal{F}_{c}$ to the goal camera frame $\mathcal{F}_{c^{*}}$ using both the image-feedback when it is available and a learned DMP when image-feedback is not available. A stable switching controller is developed, which switches between the DMP and the IBVS control. The switching is triggered when the image features are visible in the camera FOV. The feature points may or may not be visible depending on the current camera pose. In such cases, the camera motion control is switched to DMP. 

In the following sections, details of the IBVS controller and the DMP are first presented, followed by a stability analysis of the switched controller.

\section{Image-Based Visual Servo Control\label{sec:IBVS}}
The objective of IBVS is to regulate image plane feature errors to zero. Let the state vector for the current image plane features be denoted by $s_{i}(t)=\left[\begin{array}{ccc}
y_{1}(t), & \cdots & y_{2m}(t)\end{array}\right]^{T}\in\mathcal{\mathbb{R}}^{2m}$ and let the goal feature vector be denoted by $s_{i}^{*}=\left[\begin{array}{ccc}
y_{1}^{*}, & \cdots & y_{2m}^{*}\end{array}\right]^{T}$, where $m$ is the number of feature points. Each feature is represented by its $X$ and $Y$ locations in the image plane. The dynamics of the features $s_{i}(t)$ can be expressed as
\begin{equation}
\dot{s}_{i}(t)=L_{i}(s_i(t),Z(t))\xi_c(t),\label{eq:Feature-Dyn}
\end{equation}
where $L_{i}(s_i,Z)\in\mathbb{R}^{2m\times6}$ is the interaction matrix and $Z(t)=\left[\begin{array}{ccc}
Z_{1}(t) & \cdots & Z_{m}(t)\end{array}\right]^{T}\in\mathbb{R}^{m}$ is the time-varying depth of the feature points with respect to the camera frame. In (\ref{eq:Feature-Dyn}), the vector $\xi_{c}(t)=\left[\begin{array}{cc}
v_{c}^{T}(t) & \omega_{c}^{T}(t)\end{array}\right]^{T}$ contains the linear and angular camera velocities expressed in the camera frame $\mathcal{F}_{c}$ in twist coordinates.  The dependence of the variables on time and other arguments is dropped in the rest of the paper unless stated. In (\ref{eq:Feature-Dyn}), the interaction matrix is given by $L_{i}(s_{i},Z)=\left[\begin{array}{ccc}
L_{i_{1}}^{T}(y_{1},y_{2},Z_{1}) & \cdots & L_{i_{m}}^{T}(y_{2m-1},y_{2m},Z_{m})\end{array}\right]^{T}$ where 
\begin{align}L_{i_{j}}=\! & \left[\negmedspace\begin{array}{cccccc}
\frac{-1}{Z_{j}} & \!\!0 & \!\!\frac{y_{2j-1}}{Z_{j}} & \!\!y_{2j-1}y_{2j} & \!\!-(1+y_{2j-1}^{2}) & \!\!y_{2j}\\
0 & \!\!\frac{-1}{Z_{j}} & \!\!\frac{y_{2j}}{Z_{j}} & \!\!1+y_{2j}^{2} & \!\!-y_{2j-1}y_{2j} & \!\!-y_{2j-1}
\end{array}\negmedspace\right]\!\nonumber\\
 & \qquad\qquad\qquad\qquad\qquad\qquad\forall j=1,\cdots,m.\label{eq:Li}
\end{align}
The corresponding error vector between the current and the goal image features is defined as $e_{i}(t)=s_{i}(t)-s_{i}^{*}$. The error dynamics are given by
\begin{equation}
\dot{e}_{i}=L_{i}(s_{i},Z)\xi_{c}.\label{eq:IBVS_error_dyn}
\end{equation}
Taking the time derivative of (\ref{eq:IBVS_error_dyn}) yields the following second order dynamics
\begin{equation}
\ddot{e}_{i}=L_{i}(s_{i},Z)\dot{\xi}_{c}+\dot{L}_{i}(s_{i},\dot{s}_{i},Z,\dot{Z})\xi_{c}.\label{eq:IBVS_second_order}
\end{equation}
Motivated by the stability analysis, the following acceleration control law is designed,
\begin{equation}
\dot{\xi}_{c}=\widehat{L}_{i}^{+}(s_{i},Z^{\star})\left(-k_{p}e_{i}-k_{v}\dot{\hat{e}}_{i}\right),\label{eq:IBVS_control}
\end{equation}
where $k_{p},k_{v}>0$ are the controller gains, $\dot{\hat{e}}_{i}$ is an approximation of $\dot{e}_{i}(t)$, which either can be computed numerically or using an approximation of $Z(t)$ in \eqref{eq:IBVS_error_dyn}. In this paper, the depth is approximated by the constant desired depth $Z^\star$ to construct $\widehat{L}_{i}$. \\
{\textbf{Assumption 1. }}\label{assumption1} The approximation error $\chi(t) = \dot{e}(t)-\dot{\hat{e}}_{i}(t)$  is bounded by a constant, i.e., $\exists\bar{\chi}\in[0,\infty)$ such that $\sup_{t\geq0}\Vert \chi(t)\Vert\leq\bar{\chi}$.
\begin{rem}
The validity of Assumption 1 can be ensured using a suitable numerical approximation to compute $\dot{\hat{e}}_{i}$. For example, the finite difference method with smoothing or polynomial regression method over a time window can be used.  
\end{rem}
In (\ref{eq:IBVS_control}), $\widehat{L}_{i}^{+}=\left(\widehat{L}_{i}^{T}\widehat{L}_{i}\right){}^{-1}\widehat{L}_{i}^{T}$ is the pseudo-inverse of $\widehat{L}_{i}$. Using the relation $\dot{\hat{e}}_{i}(t)=\dot{e}_{i}(t)-\chi(t)$, and (\ref{eq:IBVS_error_dyn})-(\ref{eq:IBVS_control}), the resulting closed-loop second order dynamics can be written as
\begin{equation}
\ddot{e}_{i}=-k_{p}L_{i}\widehat{L}_{i}^{+}e_{i}-k_{v}L_{i}\widehat{L}_{i}^{+}\dot{e}_{i}+\dot{L}_{i}\xi_{c}+k_{v}L_{i}\widehat{L}_{i}^{+}\chi.\label{eq:IBVS_closed_loop}
\end{equation}
To facilitate the stability analysis of the switched controller, the dynamics of the state defined in \eqref{eq:State} are first established. To this end, a locally injective nonlinear function $\phi:\mathbb{R}^{6}\to\mathbb{R}^{2m}$ which maps the pose error $e_p(t)$ to image error $e_i(t)$ as  $e_{i}(t)=\phi(e_{p}(t))$ is used as described in \cite{gans2007stable}, where $\phi(0)=0$. Let $\rho(t)=\left[\begin{array}{cc}
e_{i}^{T}(t) & \dot{e_{i}}^{T}(t)\end{array}\right]^{T}\in\mathbb{R}^{4m}$. The following relation is obtained using the result in (\ref{eq:IBVS_error_dyn})
\begin{equation}
\rho =\left[\begin{array}{c}
\phi\left(e_{p}\right)\\
L_{i}\mathbf{R}\xi_{c^{*}}
\end{array}\right],\mathbf{R}=\left[\begin{array}{cc}
R_{c^{*}}^{c} & -R_{c^{*}}^{c}[t_{c}^{c^{*}}]_{\times}\\
0_{3\times3} & R_{c^{*}}^{c}
\end{array}\right].\label{eq:State_IBVS_nonlin}
\end{equation}
where $[\cdot]_{\times}$ denotes the skew symmetric operator. From \eqref{eq:State_IBVS_nonlin}, $\rho(t)=0$ when $x(t)=\left[e_p^T(t),\xi_{c^*}^T(t)\right]^{T}=0$. Using the mean value theorem and the development in Chapter 4 of \cite{Khalil2002} for (\ref{eq:State_IBVS_nonlin}) at $x(t)=0$ yields
\begin{align}
\rho & =\left[\begin{array}{cc}
J_{1} & 0_{2m\times6}\\
0_{2m\times6} & J_{2}
\end{array}\right]\left[\begin{array}{c}
e_{p}\\
\xi_{c^{*}}
\end{array}\right]+\mathcal{O}(x)\nonumber \\
\rho & =\mathbf{J}x+\mathcal{O}(x),\label{eq:IBVS_Taylor_Series}
\end{align}
where $\mathbf{J} \in \mathbb{R}^{4m\times12}$ is the Jacobian, $J_{1}=\frac{\partial\phi}{\partial e_{p}}\Big\vert_{e_{p}=0}$, $J_{2}=L_{i}\left(s_{i}^{*},Z^{\star}\right)\mathbf{R}$, the term $\mathcal{O}(x)$ satisfies $\mathcal{O}(x)\leq\upsilon_{1}\Vert x \Vert$ where $\upsilon_{1}\in\mathbb{R}^{+}$. The Jacobian $\mathbf{J}$ is full column rank in the neighborhood of $x(t)=0$ as $\phi\left(\cdot\right)$ is locally injective and $J_{2}$ can be made full column rank by selecting an appropriate desired feature vector $s_{i}^{*}$. 

To facilitate the stability analysis of the switched system, using the dynamics of $\dot{e}_{p}(t)$ derived in Section II B of \cite{gans2007stable}, the dynamics of the common state $x(t)$ for IBVS can be concisely written as a first order dynamical system of the form
\begin{equation}
\dot{x}=f_{v}\left(x,t\right)+g_{v}\left(t\right),\label{eq:VS_error_dyn}
\end{equation}where $f_{v}:\mathbb{R}^{12}\times\mathbb{R}^{+}\rightarrow\mathbb{R}^{12}$ is a locally-Lipschitz function in $x$ and piecewise continuous in $t$, whose form can be derived using the injective function $\phi$, (\ref{eq:State_IBVS_nonlin}), and (\ref{eq:IBVS_error_dyn}) and $g_{v}(t) = \left[\begin{array}{cc}
0_{1\times 6}^{T} & \left(k_{v}\mathbf{R}^{-1}\widehat{L}_{i}^{+}\chi(t)\right)^{T}\end{array}\right]^{T}$ is a perturbation term piecewise continuous term in $t$.
To facilitate the stability analysis of the system in \eqref{eq:VS_error_dyn}, following remarks and assumptions are stated. 
\begin{rem} Using the relation in (\ref{eq:IBVS_error_dyn}), the upper bound on the norm of the velocity is given by $\Vert\xi_{c}\Vert\leq\left(\inf_{t\geq0}\sigma_{\mathrm{min}}\left( L_{i}\right) \right)^{-1}\Vert\dot{e}_{i}\Vert,$ \ where $\sigma_{\mathrm{min}}\left( L_{i}\right) $ is the minimum non-zero singular value of the matrix $L_{i}(s_{i},Z)$. \end{rem}
{\textbf{Assumption 2. }}\label{assumption2}The matrices $\dot{\widehat{L}}_{i}^{+}(s_{i},Z^{\star})$ and $\dot{L}_{i}(s_{i},\dot{s}_{i},Z,\dot{Z})$ can be upper bounded using the constants $l_{1},l_{2}>0$ as $\sup_{t\geq0,\: \Vert \dot{e}_{1} \Vert  = 1}\Big\Vert\dot{\widehat{L}}_{i}^{+}\Big\Vert\leq l_{1}$ and $\sup_{t\geq 0,\: \Vert \dot{e}_{i} \Vert  = 1 }\Big\Vert\dot{L}_{i}\Big\Vert\leq l_{2}$ in a sufficiently small neighborhood of the origin.\\
\begin{rem}Assumption 2 is mild since it only requires $s_i(t),\: Z(t)$ to be bounded in a local region near $s_{i}^{*}$ (Remark 1 in \cite{rotithorTCST}). The bound on the matrix $\dot{L}_{i}(s_{i},\dot{s}_{i},Z,\dot{Z})$ holds given that $s_i(t),\: Z(t)$ are bounded, since $\dot{Z}(t)$ is a function of the bounded 3D coordinates of the feature point and the velocity vector $\xi_{c}(t)$ (Ch. 12 pp. 414  in \cite{Spong2006}), which can be upper bounded using Remark 2.\end{rem}
\textbf{Assumption 3. }\label{assumption3}The matrix $\widehat{L}_{i}^{+}$ is constructed using suitable feature points and a constant depth approximation such that in a sufficiently small neighborhood of the origin, $e_{i}(t),\dot{e}_{i}(t)\notin\mathrm{Ker}\left\{ \widehat{L}_{i}^{+}\right\}$ , $\widehat{L}_{i}^{+}L_{i}>0$  and $\widehat{L}_{i}^+$ has full column rank as described in \cite{chaumette1998potential}.
\begin{rem}
Assumption 3 implies that if the matrix $\widehat{L}_{i}^{+}$ is suitably computed and the initial camera pose is in a local neighborhood of the goal camera pose, then the feature motion is realizable which ensures the final camera pose does not reach a local minimum. In this local region the matrix $\widehat{L}_{i}^{+}L_{i}$ is positive definite \cite{chaumette1998potential}, \cite{chaumette2006visual}. 
\end{rem}
\begin{theorem}
If Assumptions 1 - 3 hold, then the dynamics in (\ref{eq:VS_error_dyn}) are uniformly ultimately bounded in the sense that for an initial time $t_{0}\geq 0$ if
\begin{equation}
\Vert x_{0}\Vert < \sqrt{\frac{\underline{\gamma_{v}}}{\overline{\gamma_{v}}}} \delta_{1} \Rightarrow \underset{t\rightarrow\infty}{\lim\sup}\Vert x\left(t\right)\Vert\leq\sqrt{\frac{\overline{\gamma_{v}}}{\underline{\gamma_{v}}}\frac{\eta}{\beta_{1}}},
\end{equation}
where $x_{0}= x(t_{0})$, provided that $\delta_{1}>\sqrt{\overline{\gamma_{v}}\eta/\underline{\gamma_{v}}\beta_{1}}$, and the controller gains $k_{p}$ and $k_{v}$ satisfy the sufficient condition
\begin{gather}
\lambda_{v}=\lambda_{\mathrm{min}}\left[\begin{array}{cc}
\frac{\varepsilon_{1}k_{p}\underline{k}}{2} & k^{\star}\\
k^{\star} & k_{v}\underline{k}-\frac{\varepsilon_{1}\bar{l}}{2}
\end{array}\right]>0,\label{eq:IBVS_gain_cond}
\end{gather}
where $k^{\star}=-\frac{\bar{l}}{2}-\frac{\bar{k}}{2}\left(k_{p}+\frac{\varepsilon_{1}k_{v}}{2}\right)$ and  $\varepsilon_{1}$, $\underline{k}$, $\bar{l}$, $\bar{k}$, $\overline{\gamma_{v}}$, $\underline{\gamma_{v}}$, $\beta_{1}$, $\eta$, $\delta_{1}$ are positive constants.
\end{theorem}
\begin{IEEEproof}
Using (\ref{eq:IBVS_Taylor_Series}), the following upper and lower bound $\underline{m}\Vert x\Vert^{2}\leq\Vert\rho\Vert^{2}\leq\bar{m}\Vert x\Vert^{2}$ can be derived where $\underline{m}=\lambda_{\mathrm{min}}\left( \mathbf{J}^{T}\mathbf{J}\right) $ and $\bar{m}=\lambda_{\mathrm{max}}\left( \mathbf{J}^{T}\mathbf{J}\right)+\bar{\upsilon}_{1}$,  where $\bar{\upsilon}_{1}\in\mathbb{R}^{+}$. The operators $\lambda_{\mathrm{min}}\left(\cdot\right) $ and $\lambda_{\mathrm{max}}\left( \cdot\right) $ correspond to the minimum nonzero and maximum eigenvalues of the matrix. Consider a candidate Lyapunov function $V_{v}(x,t):\mathcal{B}_{\delta_{1}}(0)\times \mathbb{R}^{+} \to \mathbb{R}^{+}$ such that $\mathcal{B}_{\delta_{1}}(0)\subset\mathbb{R}^{12}$ and
\begin{align}
V_{v}(x,t) & =x^{T}P_{v}(t)x + \tilde{\mathcal{O}}(x)=\rho^{T}Q(t)\rho,\label{eq:IBVS_Lyapunov}
\end{align}
with $\tilde{\mathcal{O}}(x)\leq \upsilon_{2}\Vert x \Vert^{2}$ indicating the remaining terms, where $\upsilon_{2} \in\mathbb{R}^{+}$. In  (\ref{eq:IBVS_Lyapunov}), $P_{v}(t)=\mathbf{J}^{T}Q(t)\mathbf{J}$ and $Q$ is expressed as
\begin{align}
Q=\begin{smallmatrix}\left[\begin{array}{cc}
\frac{1}{2}\left(\widehat{L}_{i}^{+}\right)^{T}\widehat{L}_{i}^{+} & \frac{\varepsilon_{1}}{4}\left(\widehat{L}_{i}^{+}\right)^{T}\widehat{L}_{i}^{+}\\
\frac{\varepsilon_{1}}{4}\left(\widehat{L}_{i}^{+}\right)^{T}\widehat{L}_{i}^{+} & \frac{1}{2}\left(\widehat{L}_{i}^{+}\right)^{T}\widehat{L}_{i}^{+}
\end{array}\right]\end{smallmatrix},
\end{align}
where $\varepsilon_{1}>0$ is a suitably chosen constant. Although $Q(t)$ is positive semidefinite, it is known from Assumption 3 that $e_{i}(t),\dot{e}_{i}(t)\notin\mathrm{Ker}\left\{ \widehat{L}_{i}^{+}\right\} $ if the initial pose of the camera is close to the goal-pose, which is enough to ensure that the quadratic form in (\ref{eq:IBVS_Lyapunov}) is always nonzero in a local region near the origin and is zero at the origin. The Lyapunov function in (\ref{eq:IBVS_Lyapunov}) can be then upper and lower bounded as $\underline{\gamma_{v}}\Vert x\Vert^{2}\leq V_{v}(x,t)\leq\overline{\gamma_{v}}\Vert x\Vert^{2}$, where $\underline{\gamma_{v}}=\inf_{t\geq0}\:\lambda_{\mathrm{min}}\left(P_{v}\right) $ and $\overline{\gamma_{v}}=\sup_{t\geq0}\:\lambda_{\mathrm{max}}\left( P_{v}\right) +\upsilon_{2}$. Taking the time derivative of (\ref{eq:IBVS_Lyapunov}), using (\ref{eq:IBVS_error_dyn}) and (\ref{eq:IBVS_closed_loop}), $\dot{V}_{v}$ can be written as
\begin{align}
\dot{V}_{v} & =e_{i}^{T}\left(\widehat{L}_{i}^{+}\right)^{T}\widehat{L}_{i}^{+}\dot{e}_{i}+\dot{e}_{i}^{T}\left(\widehat{L}_{i}^{+}\right)^{T}\widehat{L}_{i}^{+}\dot{L}_{i}\xi_{c}\nonumber\\
 & \quad\!-k_{p}\dot{e}_{i}^{T}\left(\widehat{L}_{i}^{+}\right)^{T}\widehat{L}_{i}^{+}L_{i}\widehat{L}_{i}^{+}e_{i}\negmedspace-\!k_{v}\dot{e}_{i}^{T}\left(\widehat{L}_{i}^{+}\right)^{T}\widehat{L}_{i}^{+}L_{i}\widehat{L}_{i}^{+}\dot{e}_{i}\nonumber\\
 & \quad\!+k_{v}\dot{e}_{i}^{T}\left(\widehat{L}_{i}^{+}\right)^{T}\widehat{L}_{i}^{+}L_{i}\widehat{L}_{i}^{+}\chi\enskip+\frac{\varepsilon_{1}}{2}\dot{e}_{i}^{T}\left(\widehat{L}_{i}^{+}\right)^{T}\widehat{L}_{i}^{+}\dot{e}_{i}\nonumber\\
 & \quad\!+\frac{\varepsilon_{1}}{2}e_{i}^{T}\left(\widehat{L}_{i}^{+}\right)^{T}\widehat{L}_{i}^{+}\dot{L}_{i}\xi_{c}\!-\!\frac{\varepsilon_{1}k_{p}}{2}e_{i}^{T}\left(\widehat{L}_{i}^{+}\right)^{T}\widehat{L}_{i}^{+}L_{i}\widehat{L}_{i}^{+}e_{i}\nonumber\\
 & \quad\!-\frac{\varepsilon_{1}k_{v}}{2}e_{i}^{T}\left(\widehat{L}_{i}^{+}\right)^{T}\widehat{L}_{i}^{+}L_{i}\widehat{L}_{i}^{+}\dot{e_{i}}\nonumber\\
 & \quad+\frac{\varepsilon_{1}k_{v}}{2}e_{i}^{T}\left(\widehat{L}_{i}^{+}\right)^{T}\widehat{L}_{i}^{+}L_{i}\widehat{L}_{i}^{+}\chi\nonumber\\
 & \quad\!+e_{i}^{T}\left(\widehat{L}_{i}^{+}\right)^{T}\dot{\widehat{L}}_{i}^{+}e_{i}+\dot{e}_{i}^{T}\left(\widehat{L}_{i}^{+}\right)^{T}\dot{\widehat{L}}_{i}^{+}\dot{e}_{i}\negmedspace\nonumber\\
 & \quad+\frac{\varepsilon_{1}}{2}e_{i}^{T}\left(\widehat{L}_{i}^{+}\right)^{T}\dot{\widehat{L}}_{i}^{+}\dot{e}_{i}+\frac{\varepsilon_{1}}{2}e_{i}^{T}\left(\dot{\widehat{L}}_{i}^{+}\right)^{T}\widehat{L}_{i}^{+}\dot{e}_{i}.\label{eq:Vp_dot-1}
\end{align}
Using Assumption 3, it can be ensured that $\underline{k}\Vert e_i \Vert ^2\leq e_{i}^{T}\mathrm{sym}\left\{\left(\widehat{L}_{i}^{+}\right)^{T}\widehat{L}_{i}^{+}L_{i}\widehat{L}_{i}^{+}\right\}e_{i}\leq\overline{k}\Vert e_i \Vert ^2$ and $\underline{k}\Vert \dot{e}_i \Vert ^2\leq \dot{e}_{i}^{T}\mathrm{sym}\left\{\left(\widehat{L}_{i}^{+}\right)^{T}\widehat{L}_{i}^{+}L_{i}\widehat{L}_{i}^{+}\right\}\dot{e}_{i} \leq\overline{k}\Vert \dot{e}_i \Vert ^2$ for some $\overline{k}\geq\underline{k}>0$ in the local region near the origin. Using Assumption 2 and defining the following upper bounds in the considered local region as $\sup_{t \geq 0}\big\Vert\left(\widehat{L}_{i}^{+}\right)^{T}\widehat{L}_{i}^{+}\big\Vert\leq\bar{l}$, $\left(\inf_{t\geq0}\sigma_{\mathrm{min}}\left\{ L_{i}\right\} \right)^{-1}{\sup_{t\geq0}\big\Vert\left(\!\widehat{L}_{i}^{+}\!\right)^{T}\widehat{L}_{i}^{+}\dot{L}_{i}\big\Vert\big\Vert\dot{e}_{i}\big\Vert\!\leq\!\vartheta\Vert\dot{e_{i}}\Vert^{2}}$, $\sup_{t\geq0}\Big\Vert\left(\widehat{L}_{i}^{+}\right)^{T}\dot{\widehat{L}}_{i}^{+}\Big\Vert\leq\iota\Vert\dot{e_{i}}\Vert$, $\bar{\varepsilon}=\mathrm{max}\left\{ 1,\varepsilon_{1}/2\right\} $ such that $\bar{l},\:\vartheta,\:\iota>0$. Given that the gains $k_{p}$ and $k_{v}$ are chosen according to the sufficient condition in (\ref{eq:IBVS_gain_cond}), then (\ref{eq:Vp_dot-1}) can be upper bounded as
\begin{align}
\dot{V}_{v} & \leq-\lambda_{v}\Vert e_{i}\Vert^{2}-\lambda_{v}\Vert\dot{e_{i}}\Vert^{2}+\bar{\varepsilon}\overline{k}k_{v}\bar{\chi}\left(\Vert e_{i}\Vert+\Vert\dot{e}_{i}\Vert\right)\nonumber\\
 & \quad+\iota\Vert e_{i}\Vert^{2}\Vert\dot{e}_{i}\Vert+\frac{\varepsilon_{1}}{2}\left(\vartheta+2\iota\right)\Vert e_{i}\Vert\Vert\dot{e}_{i}\Vert^{2}+\left(\iota+\vartheta\right)\Vert\dot{e}_{i}\Vert^{3}.\label{eq:Vp_dot_ub}
\end{align}
Using the inequality $\Vert e_{i}\Vert+\Vert\dot{e}_{i}\Vert\leq\sqrt{2\left(\Vert e_{i}\Vert^{2}+\Vert\dot{e}_{i}\Vert^{2}\right)}=\sqrt{2}\Vert\rho\Vert$, after some algebraic manipulations and using the bounds on $\Vert\rho\Vert$ and $\Vert x\Vert$, (\ref{eq:Vp_dot_ub}) can be simplified to
\begin{align}
\dot{V}_{v} & \leq-\frac{\beta_{1}}{\overline{\gamma}_{v}}V_{v}+\eta,\label{eq:Vp_dot_ub2}
\end{align}
$\forall\Vert\rho\Vert\leq\frac{\lambda_{v}}{2\sqrt{2}\mathrm{max}\left\{ \frac{\varepsilon_{1}}{2}\left(\vartheta+2\iota\right),\left(\iota+\vartheta\right)\right\} }$, where $\beta_{1}=\frac{\lambda_{v}\underline{m}}{4}$ and $\eta=\frac{2\overline{m}\left(\bar{\varepsilon}\overline{k}k_{v}\bar{\chi}\right)^{2}}{\lambda_{v}\underline{m}}$. The solution to the differential inequality in (\ref{eq:Vp_dot_ub2}) can be obtained using the Lemma 3.4 in \cite{Khalil2002} and is given by
\begin{align}
V_{v} \left(x,t\right) & \leq V_{v}\left(x_{0},t_{0}\right)e^{-\frac{\beta_{1}}{\overline{\gamma_{v}}}\left(t-t_{0}\right)}+\frac{\overline{\gamma_{v}}\eta}{\beta_{1}}\left(1-e^{-\frac{\beta_{1}}{\overline{\gamma_{v}}}\left(t-t_{0}\right)}\right).\label{eq:IBVS_error_bound}
\end{align}
From (\ref{eq:IBVS_error_bound}) and using the bounds on the Lyapunov function it can be concluded that $x(t)\in \mathcal{L}_{\infty}$. Using Theorem 4.18 in \cite{Khalil2002}, the state is uniformly ultimately bounded for sufficiently small $\delta_{1}>0$ representing the local region of convergence, where the relation in (\ref{eq:IBVS_Taylor_Series}), Assumptions 2-3 and (\ref{eq:Vp_dot_ub2}) holds. 
\end{IEEEproof}
The ultimate bound on the state can be made arbitrarily small by adjusting the controller gains $k_{p}$, $k_{v}$ and by choosing an appropriate approximation method for $\dot{\hat{e}}_i(t)$, which minimizes $\bar{\chi}$.
\section{Task Space Dynamic Movement Primitives\label{sec:TaskSpace-Control}}
This section describes the position and orientation DMPs which can generate goal reaching motions while retaining the desired shape of the trajectory.
\subsection{Combined Position and Orientation DMPs\label{DMPs}}
To regulate the position and orientation error to zero, i.e., $\Vert e_{p}\left(t\right)\Vert\rightarrow0$ and to retain a desired shape of the trajectory, the combined position and orientation DMPs are described by the following acceleration law
\begin{equation}
\tau^{2}\dot{\xi}_{c^{*}}=-\Gamma e_{p}-\tau\Lambda\xi_{c^{*}}+\mathbf{\Theta}^{T}\mathbf{\Psi}(z_{p},z_{o}),\label{eq:Combined_DMP_dynamics}
\end{equation}
where $\xi_{c^*}(t)$ is the velocity of the camera in the frame $\mathcal{F}_{c^*}$, $\Gamma=\mathrm{diag}\left\{ \left(\alpha_{v}\beta_{v}\right)\mathrm{I}_{3},\left(\alpha_{\omega}\beta_{\omega}\right)\mathrm{I}_{3}\right\} $, $\Lambda=\mathrm{diag}\left\{ \alpha_{v}\mathrm{I}_{3},\alpha_{\omega}\mathrm{I}_{3}\right\} $ such that $\alpha_{v},\alpha_{\omega},\beta_{v},\beta_{\omega}>0$ are constant positive gains and $\tau>0$ is the temporal scaling constant. The third term on the RHS of \eqref{eq:Combined_DMP_dynamics} is a nonlinear forcing function which encodes the shape of the desired trajectory, $\mathbf{\Theta}$ and $\mathbf{\Psi}$ are defined as
\begin{equation}
\mathrm{\mathrm{\mathbf{\Theta}}=}\left[\!\begin{array}{c}
\begin{array}{cc}
\theta_{p} & 0_{ N_{p}\times 3}\\
0_{ N_{o}\times 3} & \theta_{o}
\end{array}\end{array}\!\right],\:\text{\ensuremath{\mathbf{\Psi}}}=\left[\!\begin{array}{c}
\begin{array}{rr}
\Psi_{p}(z_{p})z_{p} \\ \Psi_{o}(z_{o})z_{o}\end{array}\end{array}\right],\label{eq:forcing_fn} 
\end{equation}
such that $\theta_{p}\in\mathbb{R}^{N_{p}\times3},\theta_{o}\in\mathbb{R}^{N_{o}\times3}$ are constant weight matrices associated with the vectors of radial basis functions $\Psi_{p}(z_{p})=\left[\frac{\psi_{1}(z_{p})}{\sum_{i=1}^{N_{p}}\psi_{i}(z_{p})},\cdots,\frac{\psi_{N_{p}}(z_{p})}{\sum_{i=1}^{N_{p}}\psi_{i}(z_{p})}\right]^{T}\in\mathbb{R}^{N_{p}}$ and $\Psi_{o}(z_{o})=\left[\frac{\varphi_{1}(z_{o})}{\sum_{i=1}^{N_{o}}\varphi_{i}(z_{o})},\cdots,\frac{\varphi_{N_{o}}(z_{o})}{\sum_{i=1}^{N_{o}}\varphi_{i}(z_{o})}\right]^{T}\in\mathbb{R}^{N_{o}}$ with $N_{p},N_{o}>0$ as respective basis numbers. The individual basis functions are defined as $\psi_{i}(z_{p})=\mathrm{exp}\left(-h_{i}^{\psi}(z_{p}(t)-c_{i})^{2}\right)$ and $\varphi_{i}(z_{o})=\mathrm{exp}\left(-h_{i}^{\varphi}(z_{o}(t)-\nu_{i})^{2}\right)$. The parameters $c_{i},\nu_{i}>0$ are centers of the basis functions and $h_{i}^{\psi},h_{i}^{\varphi}>0$ are the variances respectively. In (\ref{eq:forcing_fn}), $z_{p}\left(t\right)\in\mathbb{R}$ and $z_{o}\left(t\right)\in\mathbb{R}$ are solutions to first order scalar differential equations given by
\begin{equation}
\tau\dot{z}_{p}=-\alpha_{z_{p}}z_{p},\qquad\tau\dot{z}_{o}=-\alpha_{z_{o}}z_{o},\label{eq:Canonical System}
\end{equation}
where $\alpha_{z_{p}},\alpha_{z_{o}}>0$ are constant positive gains. To facilitate the stability analysis, the initial conditions for the differential equation in (\ref{eq:Canonical System}) are set to $z_{p}\left(t_{0}\right),z_{o}\left(t_{0}\right)=1$.\\
\textbf{Assumption 4. }\label{assumption4}The nonlinear forcing term can be upper bounded as $\Vert\mathbf{\Theta}^{T}\mathbf{\Psi}\Vert\leq\bar{\Theta}\bar{\Psi}e^{-\underline{\alpha}\left(t-t_{0}\right)}$, where $\underline{\alpha}=\frac{1}{\tau}\mathrm{min}\left\{ \alpha_{z_{p}},\alpha_{z_{o}}\right\} $ when the DMP is active between $\left[t_{0},t\right)$.
\begin{rem}
The implication of Assumption 4 is that the matrix $\mathbf{\Theta}$ is bounded by a constant, i.e., $\sqrt{\lambda_{\mathrm{max}}\left(\mathbf{\Theta}^T\mathbf{\Theta}\right)}\leq\bar{\Theta}$. This holds true since the analytical solution of the optimization problem in (\ref{eq:Optimization}) depends on the pseudo-inverse of the column-wise augmented $\mathbf{\Psi}$ matrices and the demonstration trajectories which are bounded.
\end{rem} 
Thus, based on (\ref{eq:State}), the dynamics of $\dot{e}_{p}(t)$ derived in Section II B of \cite{gans2007stable}, and (\ref{eq:Combined_DMP_dynamics}), the state dynamics for DMP can be concisely written as 
\begin{equation}
\dot{x}=f_{d}\left(x,t\right)+g_{d}\left(t\right),\label{eq:DMP_error_dyn}
\end{equation}
where $f_{d}:\mathbb{R}^{12}\times\mathbb{R}^{+}\rightarrow\mathbb{R}^{12}$ is a locally-Lipschitz function in $x$ and piecewise continuous in $t$ and $g_{d}(t) = \left[\begin{array}{cc}
0_{1\times 6}^{T} & \left(\mathbf{\Theta}^{T}\mathbf{\Psi}(z_{p},z_{o})\right)^{T}\end{array}\right]^{T}$ is a perturbation term which is piecewise continuous in $t$.
\subsection{Learning DMP Parameters}
Given a demonstration trajectory with $\mathcal{N}$ data points $\{e_{p}(t_{k}),\xi_{c^{*}}(t_{k}),\dot{\xi}_{c^{*}}(t_{k})\}_{k=0}^{\mathcal{N}-1}$ sampled at the time instants $\{t_{k}\}_{k=0}^{\mathcal{N}-1}$, the DMP weights are computed by solving the following optimization problem
\begin{align}
\mathbf{\Theta}^{\star} & =\arg\min_{\Theta}\sum_{k=0}^{\mathcal{N}-1}\Big\Vert\tau^{2}\dot{\xi}_{c^{*}}\left(t_{k}\right)+\Gamma e_{p}\left(t_{k}\right) +\tau\Lambda\xi_{c^{*}}\left(t_{k}\right) \nonumber \\ 
 & \qquad\qquad\qquad\quad-\mathbf{\Theta}^{T}\mathbf{\Psi}\left(z_{p}\left(t_{k}\right),z_{o}\left(t_{k}\right)\right)\Big\Vert^{2}, \label{eq:Optimization}
\end{align}
for suitably chosen values of $\tau$, $\Gamma$, and $\Lambda$. The centers and variances are calculated as $c_{i}=e^{-\alpha_{z_{p}}\left(\frac{i-1}{N_{p}-1}\right)}$, $h_{i}^{\psi}=\frac{1}{(c_{i+1}-c_{i})^{2}}$, $h_{N_{p}}^{\psi}=h_{N_{p}-1}^{\psi}$, $\forall i=1,\cdots,N_{p}-1$ and $\nu_{j}=e^{-\alpha_{z_{o}}\left(\frac{j-1}{N_{o}-1}\right)}$, $h_{j}^{\varphi}=\frac{1}{(\nu_{j+1}-\nu_{j})^{2}}$, $h_{N_{o}}^{\varphi}=h^{\varphi}_{N_{o}-1}$, $\forall j=1,\cdots,N_{o}-1$. 

\subsection{Stability Analysis of DMPs}
This subsection presents the stability analysis of the DMP described in Section \ref{DMPs}.\\
 \textbf{Property 1.}\label{property1} \cite{han2008control} Based on the definition of $r\left(t\right)$ defined after (\ref{eq:State}), the following relations hold,
\begin{equation}
r^{T}\dot{r}=r^{T}\omega_{c^{*}},\quad\dot{r}^{T}\omega_{c^{*}}\leq\omega_{c^{*}}^{T}\omega_{c^{*}}.\label{eq:angle-axis-identity}
\end{equation}
\begin{theorem}
Provided that Assumption 4 holds, then the system in (\ref{eq:DMP_error_dyn}) is globally asymptotically stable in the sense that 
\begin{align}
\lim_{t\rightarrow \infty}\Vert x(t) \Vert = 0,
\end{align} 
given that the gains $\alpha_{v}$,  $\beta_{v}$, $\alpha_{\omega}$ and $\beta_{\omega}$ are chosen according to the sufficient condition
\begin{equation}
\alpha_{v}=4\beta_{v},\:\alpha_{\omega}=4\beta_{\omega},\:\beta_{v}>\frac{3\varepsilon_{2}}{8\tau},\:\beta_{\omega}>\frac{3\varepsilon_{2}}{8\tau},\label{eq:DMP_gain_cond}
\end{equation}
where $\varepsilon_{2}\in\left(0,\sqrt{4\tau^{2}\lambda_{\min}\left(\Gamma\right)}\right)$ is a positive constant.
\end{theorem}
\begin{IEEEproof}
Consider the candidate Lyapunov function $V_{d}(x):\mathbb{R}^{12}\to\mathbb{R}^{+}$ defined as
\begin{equation}
V_{d}(x)=x^{T}P_{d}x,\label{eq:DMP_Lyapunov}
\end{equation}
where $P_{d}=\left[\begin{array}{cc}
\frac{1}{2}\mathrm{\Gamma} & \frac{\varepsilon_{2}}{4}\mathrm{I_{6}}\\
\frac{\varepsilon_{2}}{4}\mathrm{I_{6}} & \frac{\tau^{2}}{2}\mathrm{I}_{6}
\end{array}\right]$. The Lyapunov function in (\ref{eq:DMP_Lyapunov}) is upper and lower bounded as $\underline{\gamma_{d}}\Vert x\Vert^{2}\leq V_{d}(x)\leq\overline{\gamma_{d}}\Vert x\Vert^{2}$, where $\underline{\gamma_{d}}=\lambda_{\mathrm{min}}\left(P_{d}\right) $ and $\overline{\gamma_{d}}=\lambda_{\mathrm{max}}\left(P_{d}\right) $. Taking the time derivative of $V_{d}\left(x\right)$, using the definition of $x(t)$ in  (\ref{eq:State}), substituting (\ref{eq:angle-axis-identity}), (\ref{eq:Combined_DMP_dynamics}), and simplifying yields
\begin{align}
\dot{V}_{d} & \leq-\tau\xi_{c^{*}}^{T}\Lambda\xi_{c^{*}}+\xi_{c^{*}}^{T}\Theta^{T}\Psi-\frac{\varepsilon_{2}}{2\tau^{2}}e_{p}^{T}\Gamma e_{p}-\frac{\varepsilon_{2}}{2\tau}e_{p}^{T}\Lambda\xi_{c^{*}}\nonumber \\
 & \quad+\frac{\varepsilon_{2}}{2\tau^{2}}e_{p}^{T}\Theta^{T}\Psi+\frac{\varepsilon_{2}}{2}\xi_{c^{*}}^{T}\xi_{c^{*}}.\label{eq:DMP_Lyapunov_deriv2}
\end{align}
The DMP gain relations are selected as per \cite{ude2014orientation} to be $\alpha_{v}=4\beta_{v}$ and $\alpha_{\omega}=4\beta_{\omega}$ to make a proportional-derivative like part of the system critically damped. Defining the constants $\bar{c}_{1}\geq\frac{\varepsilon_{2}\bar{\Theta}\overline{\Psi}}{2\tau^{2}}$, $\bar{c}_{2}\geq\frac{2\tau^{2}\bar{c}_{1}}{\varepsilon_{2}}$, $C=\mathrm{max}\left\{ \bar{c}_{1},\bar{c}_{2}\right\} $, if $\beta_v$ and $\beta_{w}$ are chosen according to the sufficient conditions in (\ref{eq:DMP_gain_cond}), then (\ref{eq:DMP_Lyapunov_deriv2}) can be upper bounded as $\dot{V}_{d}\leq-\lambda_{d}\Vert x\Vert^{2}+\sqrt{2}Ce^{-\underline{\alpha}\left(t-t_{0}\right)}\Vert x\Vert$ where $\lambda_{d}=\mathrm{min}\left\{ \frac{\varepsilon_{2}\beta_{v}^{2}}{\tau^{2}},4\tau\beta_{v}-\frac{3\varepsilon_{2}}{2},\frac{\varepsilon_{2}\beta_{\omega}^{2}}{\tau^{2}},4\tau\beta_{\omega}-\frac{3\varepsilon_{2}}{2}\right\} $. Completing the squares and using the bounds on the Lyapunov function yields
\begin{align}
\dot{V}_{d} & \leq-\frac{\lambda_{d}}{2\bar{\gamma}_{d}}V_{d}+\frac{C^{2}}{\lambda_{d}}e^{-2\underline{\alpha}\left(t-t_{0}\right)}\nonumber \\
 & \leq-\beta_{2}V_{d}+\varpi e^{-2\underline{\alpha}\left(t-t_{0}\right)},\label{eq:DMP_Lyapunov_deriv5}
\end{align}
where $\beta_{2}=\frac{\lambda_{d}}{2\bar{\gamma}_{d}},\:\varpi=\frac{C^{2}}{\lambda_{d}}$. The solution to the differential inequality in (\ref{eq:DMP_Lyapunov_deriv5}) can be obtained using Lemma 3.4 in \cite{Khalil2002} as
\begin{equation}
V_{d}\left(x(t)\right)\leq V_{d}\left(x(t_{0})\right)e^{-\beta_{2}\left(t-t_{0}\right)}+\bar{\varrho}e^{-\underline{c}_{3}\left(t-t_{0}\right)},\label{eq:Vd_soln}
\end{equation}
where $\underline{c}_{3}=\mathrm{min}\left\{ 2\underline{\alpha},\beta_{2}\right\} $, $\bar{\varrho}=\frac{\varpi}{\vert\beta_{2}-2\underline{\alpha}\vert}$ and  $\beta_{2}\neq 2\underline{\alpha}$ such that $\vert\cdot\vert$ denotes the absolute value of the argument. Using (\ref{eq:Vd_soln}) and the bounds on the Lyapunov function, it can be concluded that $\Vert x(t) \Vert \in \mathcal{L}_{\infty}$. The second term on the right hand side of (\ref{eq:DMP_Lyapunov_deriv5}) is a non-negative continuous perturbation term which satisfies $\underset{t\rightarrow\infty}{\lim}\varpi e^{-2\underline{\alpha}\left(t-t_{0}\right)}=0$ resulting in the global asymptotic stability of the solution trajectories $x(t)$ (c.f. Lemma 9.6 in \cite{Khalil2002}). 
\end{IEEEproof}
\section{Stability Analysis of the Switched System\label{sec:Hybrid-System}}
In this section, the stability of switched system is analyzed while switching between DMP and IBVS controllers. To facilitate the stability analysis the common state defined in \eqref{eq:State} and stability results, which are derived for the dynamics of the common state using IBVS control and DMP, are used. The switched system switches between the IBVS controller in (\ref{eq:IBVS_control}) and DMP in (\ref{eq:Combined_DMP_dynamics}) based on the visibility of feature points. The switching between the controllers leads to discontinuous dynamics which can be mathematically expressed  as
\begin{equation}
\dot{x}=f_{\sigma\left(x,t\right)}\left(x,t\right)+g_{\sigma(x,t)}(t),\:\:\:\sigma:\mathbb{R}^{12}\times\mathbb{R}^{+}\rightarrow \left\{ v,d\right\} .\label{eq:Switched_dynamics}
\end{equation}
The switching signal $\sigma(x,t)$ governs the dynamics of $x(t)$. Since the IBVS exponentially converges to a bound when initialized in a local region near the goal state, a threshold constant $\delta_{2}>0$ is selected for the IBVS controller to be active such that, $x(t)\in\mathcal{B}_{\delta_{2}}(0)\subset\mathcal{B}_{\delta_{1}}(0)$. The relation between the positive constants $\delta_{1}$ and $\delta_{2}$ is established using the analysis in Theorem 3. 
\begin{rem}
In practice, a threshold for image pixel error $e_{i}(t)$ and velocity $\xi_{c^*}(t)$, which corresponds to $\delta_{2}$, can be chosen based on the image size and feature visibility. The IBVS system is active when all the features are detected, matched and $\underline{\iota}\leq e_{i}(t)\leq \overline{\iota}$ for some suitably chosen vector $\overline{\iota}>\underline{\iota}$, where the inequalities are element-wise. No threshold is selected for the velocities $\xi_{c^*}(t)$, however, the constant $\tau$ in (\ref{eq:Combined_DMP_dynamics}) can be adjusted to generate slow velocities using the DMP. 
\end{rem}
\begin{figure}[h]
\begin{centering}
\includegraphics[width=1\columnwidth]{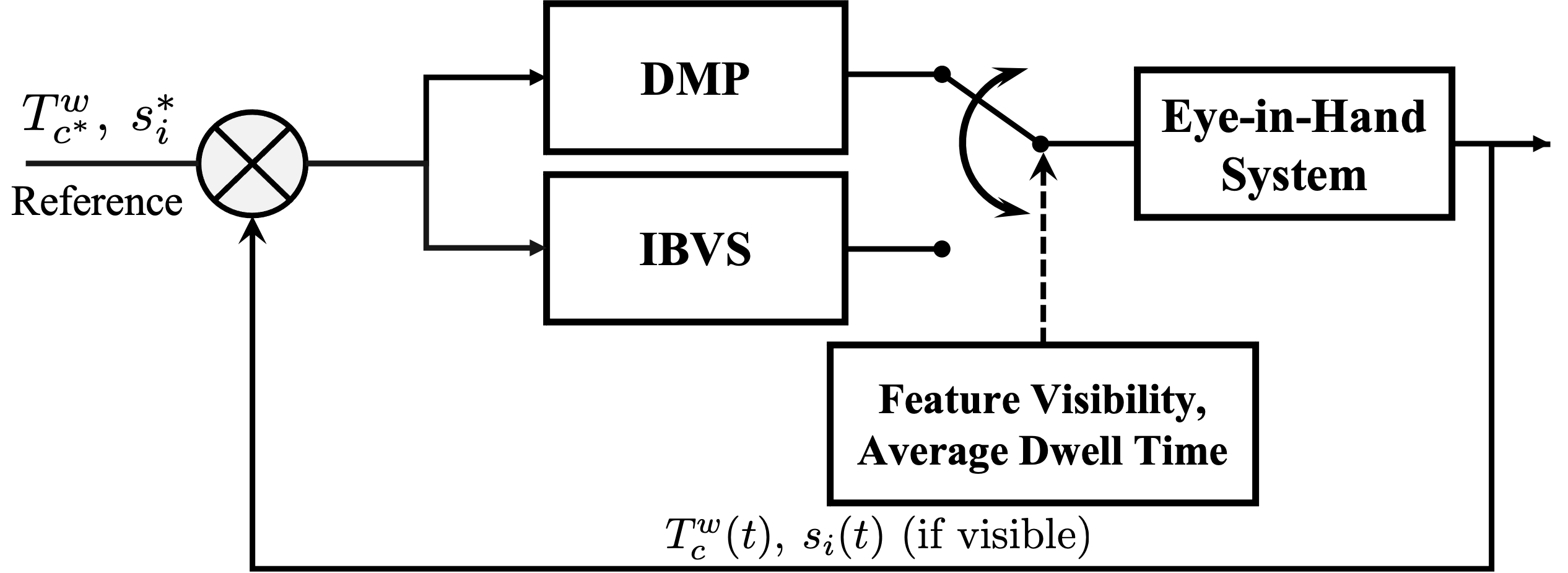}
\par\end{centering}
\begin{centering}
\caption{\label{fig:SwitchingBD}Block diagram of the DMP-IBVS switched system.}
\par\end{centering}
\end{figure}
Let $\bar{\kappa}=\mathrm{max}\left\{ \overline{\gamma_{v}},\overline{\gamma_{d}}\right\}, \:\underline{\kappa}=\mathrm{min}\left\{ \underline{\gamma_{v}},\underline{\gamma_{d}}\right\}, \:\mu=\frac{\bar{\kappa}}{\underline{\kappa}}$ be constants. The Lyapunov functions $V_v(x,t)$ and $V_d(x)$ in (\ref{eq:IBVS_Lyapunov}) and (\ref{eq:DMP_Lyapunov}) can then be related as follows.
\begin{equation}
V_{v}\leq\mu V_{d},\quad V_{d}\leq\mu V_{v},\quad\forall x(t)\in\mathcal{B}_{\delta_{1}}(0).\label{eq:Multiple_Lyapunov_Functions}
\end{equation}
\begin{definition} (Ch. 3, pp. 58 in \cite{liberzon2003switching})
The switching signal $\sigma(x,t)$ has average dwell time $\tau_{a}$, if there exist two numbers $N_{0}\in\mathbb{Z}^{+}$ and $\tau_{a}\in\mathbb{R}^{+}$ such that
\begin{equation}
N_{\sigma}\left(t,\underline{t}\right)\leq N_{0}+\frac{t-\underline{t}}{\tau_{a}}\label{eq:Dwell_time_condition}
\end{equation}
is satisfied, where $N_{0}$ is known as the chatter bound and $N_{\sigma}(t,\underline{t})$ are the number of discontinuities on the interval $[\underline{t},t)$.
\end{definition}
\begin{theorem}
Provided that Assumptions 1 - 4 hold, the state trajectories of the switched system generated by the family of subsystems described by (\ref{eq:Switched_dynamics}) and a piecewise constant, right continuous switching signal $\sigma:\mathbb{R}^{12}\times\mathbb{R}^{+}\rightarrow \left\{ v,d\right\} $ asymptotically converge to a bound in the sense that
\begin{align}
\underset{t\rightarrow\infty}{\lim\sup}\Vert x\left(t\right)\Vert\leq\sqrt{\frac{\overline{\kappa}^{N_{0}+1}\eta}{\underline{\kappa}^{{N_{0}}+2}\epsilon}} \label{eq:Switched-bound}
\end{align}provided that\begin{enumerate}[I.]
\item If $x(t)\in\mathbb{R}^{12}\setminus\mathcal{B}_{\delta_{1}}(0) \Rightarrow \sigma(x,t)=d$. 
\item If $\sigma(x(t^{-}),t^{-})=d$ and $x(t)\in \mathcal{B}_{\delta_{2}}\left(0\right)$, then $\sigma(x(t),t)=v$ .
\item The local region of convergence and the switching threshold satisfy the following conditions \begin{align}
			\sqrt{\frac{\underline{\kappa}^{N_{0}+2}}{\overline{\kappa}^{N_{0}+2}}}\delta_{1}>\delta_{2},\:\mathrm{and}\: \delta_{1}>\sqrt{\frac{\overline{\kappa}^{N_{0}+1}\eta}{\underline{\kappa}^{{N_{0}}+2}\epsilon}} \label{eq:Condition2}.
\end{align}
\item The condition in (\ref{eq:Dwell_time_condition}) is satisfied for
\begin{align}
\tau_{a}>\frac{\mathrm{ln}\:\mu}{\underline{\beta}-\epsilon}, \label{eq:time_cond}
\end{align}
\end{enumerate}where $0<\epsilon<\underline{\beta}\leq\mathrm{min}\left\{ \frac{\beta_{1}}{\overline{\gamma_{v}}},\underline{c}_{3}\right\} $.
\end{theorem}
\begin{IEEEproof}
The proof of the theorem is divided into two cases to characterize the behavior of the switched system in different regions of the space. Consider the Lyapunov function defined in (\ref{eq:IBVS_Lyapunov}) and (\ref{eq:DMP_Lyapunov})
\begin{equation}
V_{\sigma(x,t)}(x,t)=\begin{cases}
x^{T}P_{v}(t)x+\tilde{\mathcal{O}} & \sigma\left(x,t\right)=v\\
x^{T}P_{d}x & \sigma\left(x,t\right)=d
\end{cases}. \label{eq:Switched_Lyapunov}
\end{equation}
\textbf{Case 1.\:}When $\Vert x(t)\Vert>\delta_{2}$, the state $x\left(t\right)\in\mathbb{R}^{12}\setminus\mathcal{B}_{\delta_{2}}(0)$ is outside of the switching set for IBVS. If Condition I of Theorem 3 is satisfied, then $V_{\sigma(x,t)}\left(x,t\right)$ is differentiable and decreases exponentially from any initial condition. Then $\forall t\in\left[0,t_{0}\right)$ for $t_{0}>0$, the following upper bound holds
\begin{align}
V_{\sigma}\left(x(t),t\right) & \leq V_{\sigma}\left(x(0),0\right)e^{-\beta_{2}t}+\bar{\varrho}e^{-\underline{c}_{3}t}\nonumber\\
 & \leq\left(V_{\sigma}\left(x(0),0\right)+\bar{\varrho}\right)e^{-\underline{\beta} t},\label{eq:thm3_vdot1}
\end{align}
where $\sigma:\mathbb{R}^{12}\setminus\mathcal{B}_{\delta_{2}}\left(0\right)\times\left[0,t_{0}\right)\rightarrow d$ and the arguments of $\sigma$ are dropped in (\ref{eq:thm3_vdot1}) for brevity. The time required to reach the set $\mathcal{B}_{\delta_{2}}(0)$ from any initial condition is finite and can be lower bounded by $ t_{0}\geq\frac{1}{\underline{\beta}}\mathrm{ln}\left(\mathrm{min}\left\{ \frac{\overline{\gamma_{d}}\Vert x\left(t_{0}\right)\Vert^{2}+\bar{\varrho}}{\underline{\gamma_{d}}\delta_{2}^{2}},1\right\} \right)$. The controller  switches to IBVS when $\Vert x(t_{0})\Vert<\delta_{2}$. The switched system state trajectories after switching to and operating in IBVS thereafter remain in a subset of the local region, i.e., $\Vert x(t_{0})\Vert < \delta_{2} \Rightarrow\Vert x(t)\Vert<\delta_{1}$, if $\sigma(x,t)=v,\:\forall\:t\geq t_{0}$, and condition III is satisfied.

\textbf{Case 2.\:}Let $t_{0}=\inf\left\{ t\in\mathbb{R}^{+}\big\vert\Vert x\left(t\right)\Vert<\delta_{2}\right\}\geq0$ be the first time instance when $\Vert x(t)\Vert<\delta_{2}$ implying that $x\left(t_{0}\right)\in\mathcal{B}_{\delta_{2}}(0)$. Such a $t_{0}<\infty$ exists due to exponentially decaying upper bound in (\ref{eq:thm3_vdot1}). Using a recursion similar to the results in \cite{vu2007input}, Lyapunov bounds in (\ref{eq:IBVS_error_bound}), (\ref{eq:Vd_soln}), and the average dwell time condition in (\ref{eq:Dwell_time_condition}) and (\ref{eq:time_cond}), the conservative bound on the solution of the Lyapunov function in (\ref{eq:Switched_Lyapunov}) between the interval $\left[t_{0},t\right)$ for monotonically increasing sequence of switching times $\left\{ t_{n}\right\} _{n=0}^{N_{\sigma}\left(t_{0},t\right)}$ can be given as 
\begin{align}
V_{\sigma}\left(x(t),t\right)&\!\leq\mu^{N_{0}+1}\!\left(\left(V_{\sigma}\left(x(t_{0}),t_{0}\right)\!+\!\bar{w}\right)e^{-\epsilon\left(t-t_{0}\right)}\right. \nonumber\\
&\left.\:+\frac{\eta}{\epsilon}\left(1-e^{-\epsilon\left(t-t_{0}\right)}\right)\right),\quad\forall x(t)\in\mathcal{B}_{\delta_{1}}(0)\label{eq:Switched_Lyapunov_soln}
\end{align} where $\bar{w}=\varpi/\left(2\underline{\alpha}-\epsilon\right)
$. For large enough $t$, the solution $x(t)$ never exits the region $\mathcal{B}_{\delta_{1}}(0)$. It can be concluded from (\ref{eq:Switched_Lyapunov_soln}) that $x(t)$ is continuous, $x(t)\in \mathcal{L}_{\infty}$ and converges asymptotically to the bound in (\ref{eq:Switched-bound}).
\end{IEEEproof}
The ultimate bound on the switched system is larger compared to the bound on the individual IBVS system as switching between subsystems can lead to growth of the Lyapunov function. Such a bound is commonly found in the literature in different contexts (c.f. Theorem 1 in \cite{klotz2018}, Lemma 2 in \cite{veer2019switched}, Theorem 1 in \cite{ye2020switching}).
Algorithm 1 is developed based on the stability analysis presented in Theorem 3 to ensure stable switching by compensating for the fast switching occurring due to the disturbances and occlusions in the IBVS subsystem. A compensation time $t_{c}$  is calculated and the DMP subsystem is kept active for additional time to ensure that the average dwell time condition is satisfied for a total of $\bar{N}>0$ switches.
\begin{rem}
Theorem 3 establishes the general conditions for switching between the subsystems and is useful in the case of frequent perturbations or occlusions which can result in fast switching which can make the combined system unstable if the average dwell time condition is not satisfied. 
\end{rem}
\begin{rem}
The constant $\underline{\beta}$ depends on the IBVS controller gains $k_{p}$, $k_{v}$ and the DMP gains $\alpha_{v}$, $\alpha_{\omega}$, $\beta_{v}$, $\beta_{\omega}$, $\alpha_{z_{p}}$, $\alpha_{z_{o}}$. Increasing the gains can increase $\underline{\beta}$ which enables the user to choose $\epsilon$ from a wider range of values. Choosing a high value of $\epsilon$ leads to a faster convergence of the bound in (\ref{eq:Switched_Lyapunov_soln}). However, this increases the average dwell time $\tau_{a}$ in (\ref{eq:time_cond}). On the other hand, if $\epsilon$ is chosen to be  small, the bound in (\ref{eq:Switched_Lyapunov_soln}) converges slowly but allows for a lower $\tau_{a}$.
\end{rem}

\begin{algorithm}[t]
Select decay rates $\underline{\beta},\epsilon>0$, chatter bound $N_0\geq1$, feature error thresholds $\overline{\iota}>\underline{\iota}$\;
Compute $\mu = \frac{\overline{\kappa}}{\underline{\kappa}}$ and $\tau_{a}=\frac{\ln\:\mu}{\underline{\beta}-\epsilon}$\;
Set $t_d=0$ and $t_{v}=0$ to be the last time instant when DMP and IBVS were active respectively\;
Set $t_e=0$ as the elapsed time and $\bar{N}>1$  as the number of switches over which the dwell time condition should be satisfied\;
Set the compensation time $t_{c}=0$ and current number of switches $N_{\sigma}=0$\;
\While{data is available}{
\uIf {features match $\&$ $\underline{\iota}\leq e_{i} \leq \overline{\iota}$ $\&$ $t>t_v+t_c$}{
Run IBVS according to (\ref{eq:IBVS_control})\;
\uIf {$t_d>t_v$}{
$t_{e} \leftarrow t_{e} + (t_{d} - t_{v})$\;
$N_{\sigma} \leftarrow N_{\sigma }+ 1$\;
$t_{c}\leftarrow 0$\;
}
$t_{v}\leftarrow t$\;
}
\uElse{
Run DMP according to (\ref{eq:Combined_DMP_dynamics})\;
\uIf{$t_{v}>t_{d}$}{
$N_{\sigma} \leftarrow N_{\sigma }+ 1$\;
$t_{e} \leftarrow t_{e} + (t_{v} - t_{d})$\;
\uIf{$N_{\sigma}\geq\bar{N}$}{
$t_c\leftarrow (N_{\sigma}-N_{0})\tau_{a} - t_{e}$\;
$t_{e}\leftarrow 0$\;
$N_\sigma\leftarrow 0 $\;}
}
$t_d\leftarrow t$\;
}
}
\caption{Algorithm for switching between IBVS and DMP based on visibility of feature points and average dwell time condition}
\label{I-Algorithm} 
\end{algorithm}

\section{Experiments\label{sec:Experiments}}

\subsection{Experimental Setup}
\begin{figure}
    \centering
    {\includegraphics[width=0.9\columnwidth]{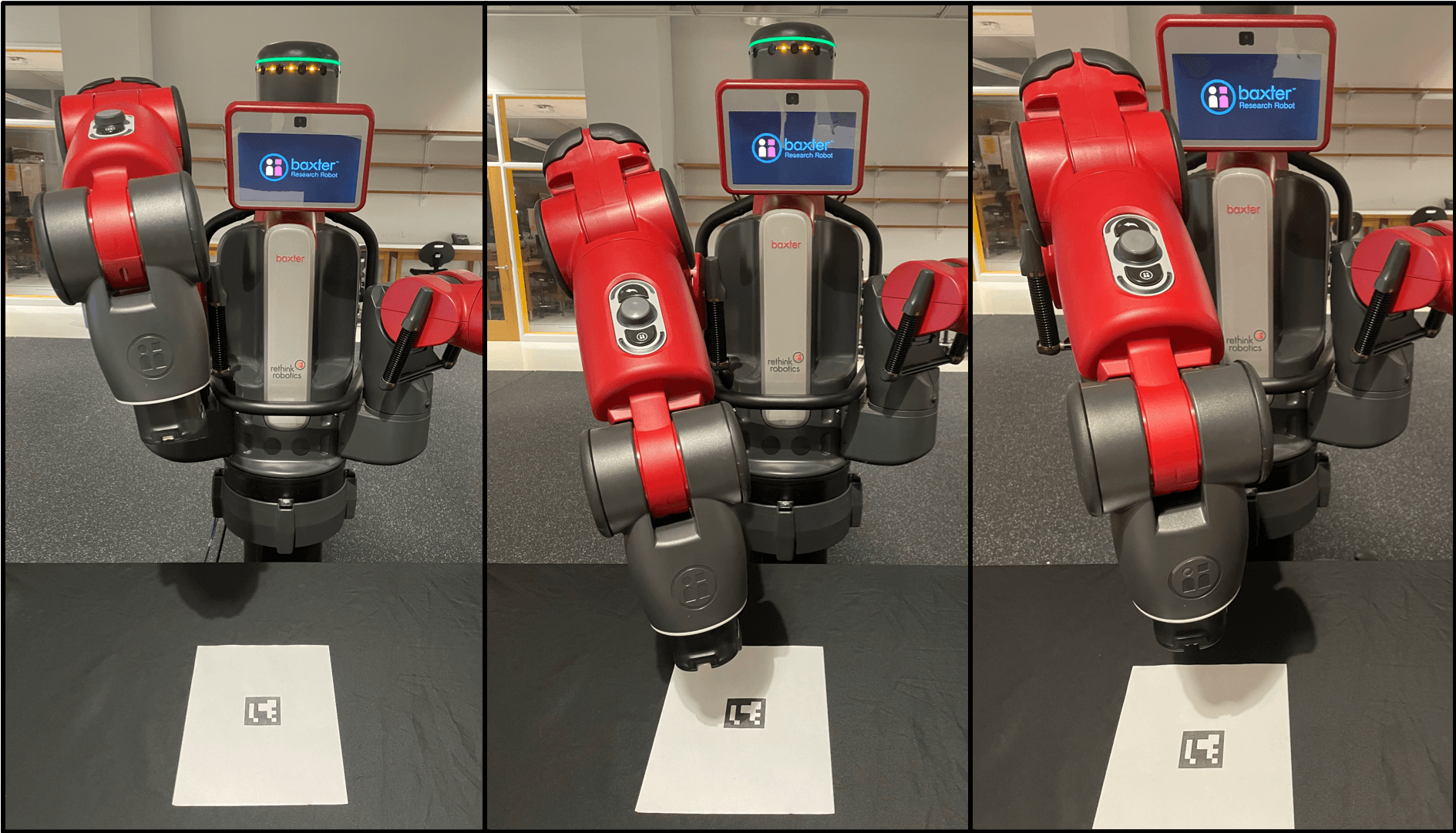}}
    \caption{The experimental setup showing Baxter's end-effector with an eye-in-hand configuration observing the ArUco marker.}
    \label{fig:baxter}
\end{figure}
The camera in the wrist of the right arm of a Baxter research robot is used to capture images of an ArUco marker with the frame rate of
30 fps and image resolution $640\times420$. The corners of the ArUco marker are used as feature points ($m=4$) for IBVS. The feature points are detected and matched using OpenCV's ArUco library. The processing is done at $30\:\mathrm{fps}$ using a desktop with an Intel Core2Duo CPU with clock-speed of 2.26 GHz and 4 GB RAM running Ubuntu 14.04. The experimental setup is shown in Fig. \ref{fig:baxter}. The DMP-IBVS switching algorithm is implemented in MATLAB 2018a. The camera intrinsic parameters for Baxter's right hand camera are given by $f_{x}=f_{y}=407.1$, $c_{x}=323.4$ and $c_{y}=205.6$, where $f_{x},\:f_{y}$  represent the camera focal lengths in pixels and $\left(c_{x},c_{y}\right)$ represents the camera center pixel.

The position DMP is trained on a demonstration trajectory, which is collected by recording Baxter's arm position, velocity, acceleration and filtered using a constant acceleration Kalman filter. The orientation DMP is trained on a trajectory generated by a minimum-jerk polynomial for quaternions. The goal-pose of the DMP is used to record the desired feature vector $s_{i}^{*}$ for IBVS system. The DMP acceleration commands are computed from the Baxter robot pose feedback. The IBVS and DMP controller-generated acceleration is integrated and applied as a joint velocity controller using the pseudo-inverse of the Baxter manipulator Jacobian matrix.

\subsection{Experiment 1}
\begin{figure*}
\begin{centering}
\makebox[1\linewidth][c]{\subfigure[]{\includegraphics[width=0.22\paperwidth]{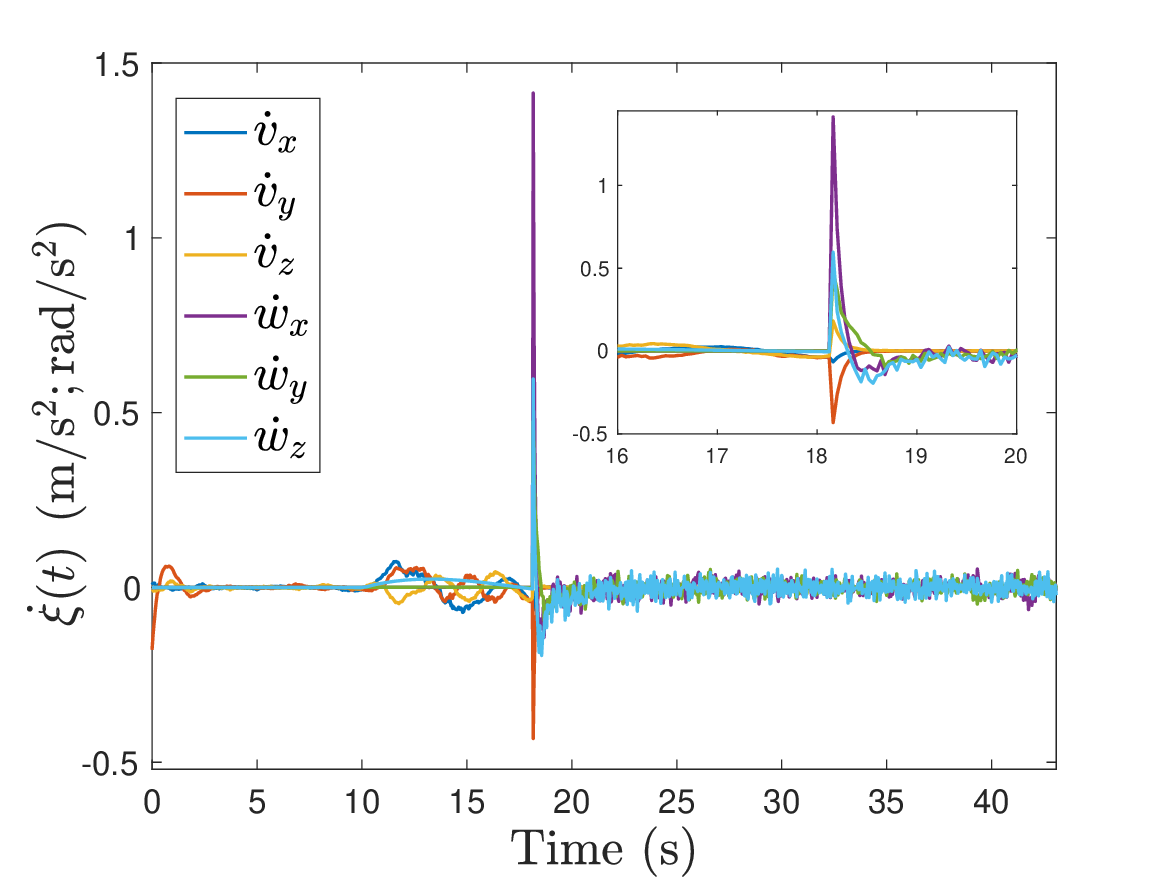}}\hspace{-0.5em}\subfigure[]{\includegraphics[width=0.22\paperwidth]{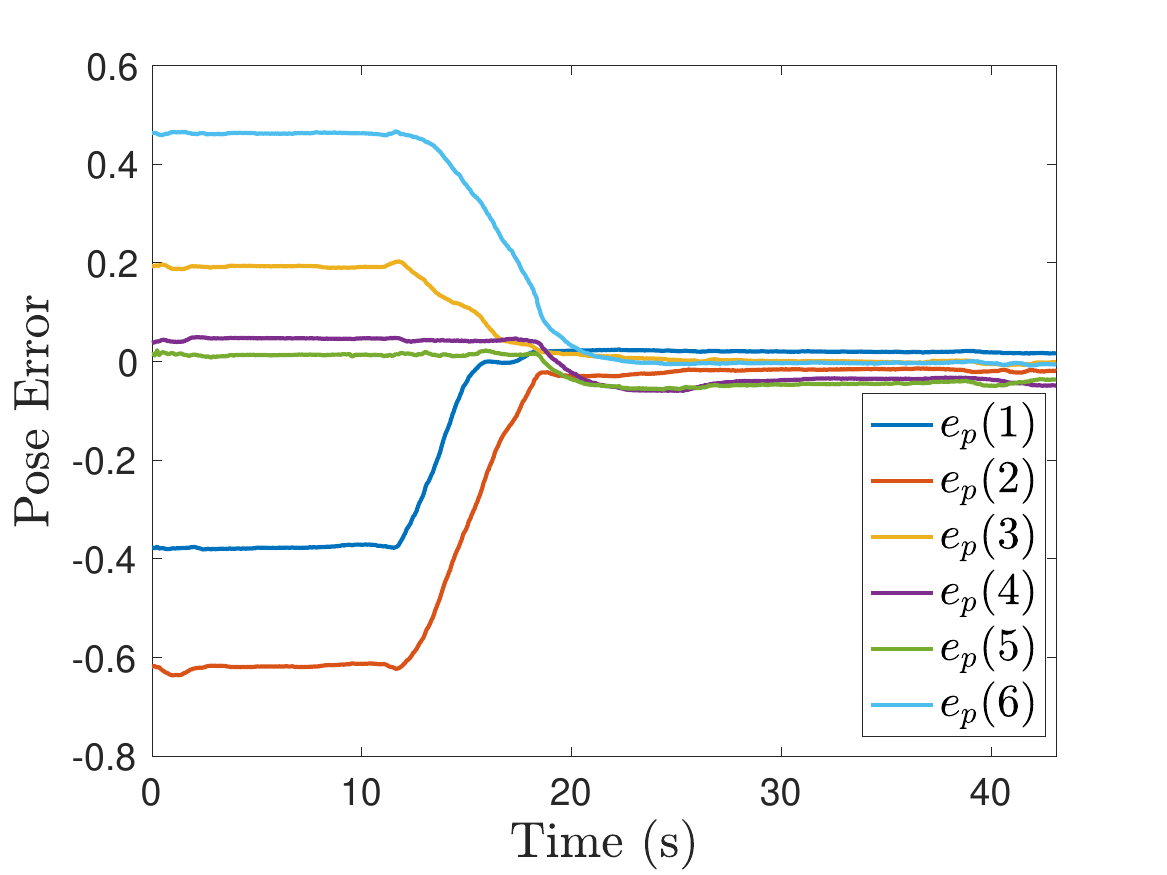}}\hspace{-0.5em}\subfigure[]{\includegraphics[width=0.22\paperwidth]{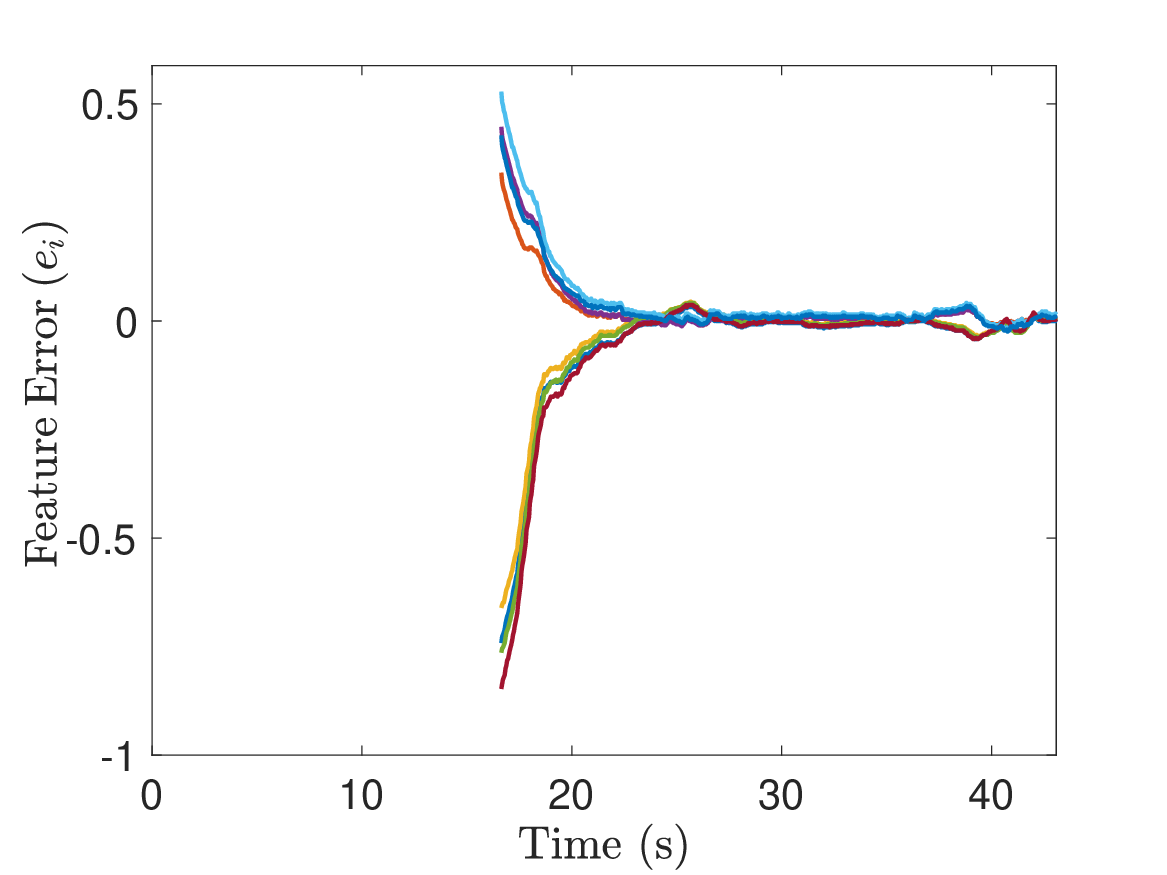}}\hspace{-0.5em}\subfigure[]{\includegraphics[width=0.22\paperwidth]{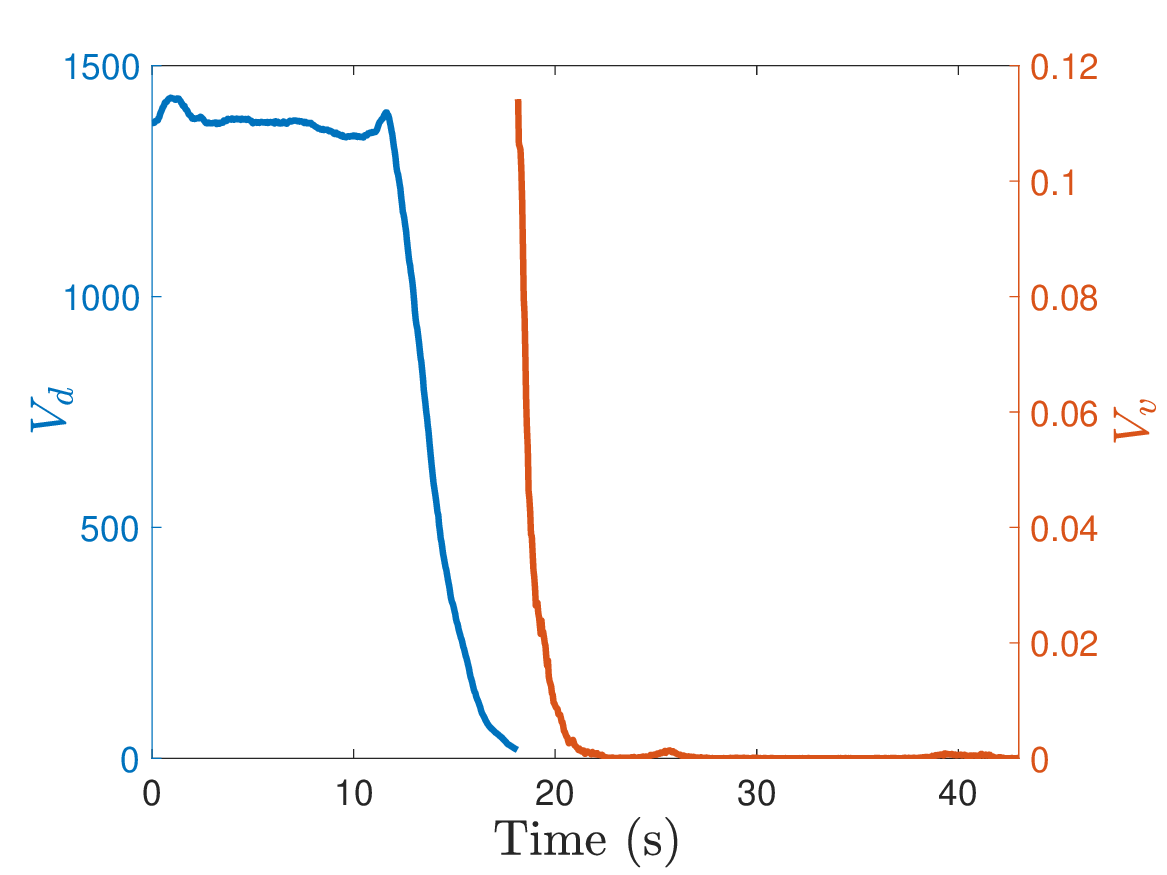}}}\\\vspace{-1.5ex}
\par\end{centering}
\centering{}\caption{Experimental results for the DMP and IBVS for a single switching instance at $16.4\:\mathrm{s}$. (a) Camera acceleration generated by DMP controller till $16.4\:\mathrm{s}$ and camera acceleration generated by IBVS controller after $16.4\:\mathrm{s}$  (b) Pose error $e_{p}(t)$ converges to a bound.
(c) Image feature errors $e_{i}(t)$ computed after $16.4\:\mathrm{s}$. (d) Value of Lyapunov function $V_{\sigma(x,t)}(x(t),t)$, with left y-axis showing the scale for $V_{d}(x(t))$ and right y-axis showing the scale for $V_{v}(x(t),t)$.\label{Experiment1}}
\end{figure*}
In this experiment, the convergence and stability of the DMP-IBVS switched system is verified when switching occurs only once. The DMP weights are computed by recording a trajectory as described earlier. The DMP parameters are set to $\alpha_{v} = 140$, $\beta_{v}=35$, $\alpha_{\omega} = 4$, $\beta_\omega=1$, $\alpha_{z_{p}}=1$, $\alpha_{z_{o}}=1$, and $\tau = 25$. The IBVS gains are empirically tuned to be $k_{p} = 5$, and $k_{v} = 10$. The initial pose of the Baxter right-hand camera is selected such that none of the desired feature points are in the FOV of the camera. The feature threshold error for switching is chosen as $\overline{\iota} = \left[0.85,0.42,\:\cdots\:,0.85,0.42\right]^T$, $\underline{\iota}=-\overline{\iota}$ which corresponds to pixel errors of $345$ pixels and $170$ pixels in the $X$ and $Y$-directions of the image frame, respectively. When the features are out of the FOV the arm starts approaching the goal location along the trained DMP trajectory. The controller switches to IBVS once the features are in the FOV of the camera and the feature error is less than the selected threshold error, i.e, $\underline{\iota}\leq e_{i}\leq\overline{\iota}$. This condition is first satisfied at $t=16.4\:\mathrm{s}$. Fig. \ref{Experiment1}(a) shows the acceleration generated by the switched system with the discontinuity at $16.4\:\mathrm{s}$, which is the switching instance. The pose error starts converging to the desired pose as the DMP is active until $t=16.4\:\mathrm{s}$. Once the switch to IBVS happens, the pose error in Fig. \ref{Experiment1}(b) converges to a bound. It is also observed that the pose error is continuous as proven in Theorem 3. Fig. \ref{Experiment1}(c) shows the image feature error $e_{i}(t)$, which is first computed when all the features are visible and the threshold condition is satisfied. The image feature error decreases exponentially to a bound as the IBVS system converges. Fig. \ref{Experiment1}(d) shows the Lyapunov function $V_{\sigma(x,t)}(x(t),t)$ for the switched system. The orange line shows $V_{v}(x(t),t)$ and the right y-axis represents its values. Similarly, the blue line shows $V_{d}(x(t))$ and the left y-axis represents its values. The Lyapunov functions are only plotted for the active subsystem in the corresponding time interval. The Lyapunov function asymptotically decreases for both controllers when active individually, verifying the result of Theorem 3. The error bounds obtained for the experiment in Table \ref{tab:error_bound} verify the convergence of the error for the switched system. The error bounds are computed by taking the average of the last 5 error data points.

\subsection{Experiment 2}
\begin{figure*}
\begin{centering}
\makebox[1\linewidth][c]{\subfigure[]{\includegraphics[width=0.22\paperwidth]{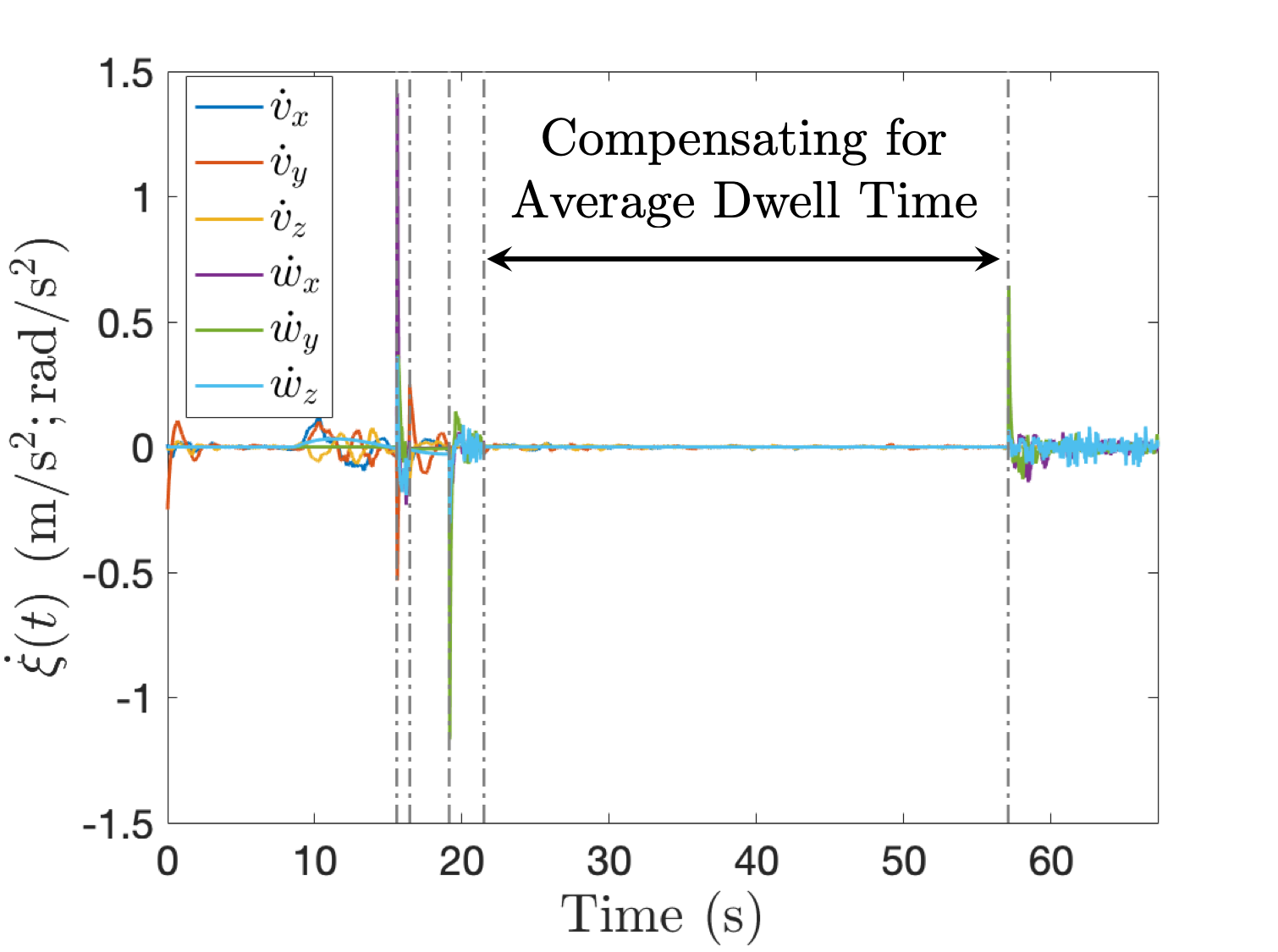}}\hspace{-0.75em}\subfigure[]{\includegraphics[width=0.22\paperwidth]{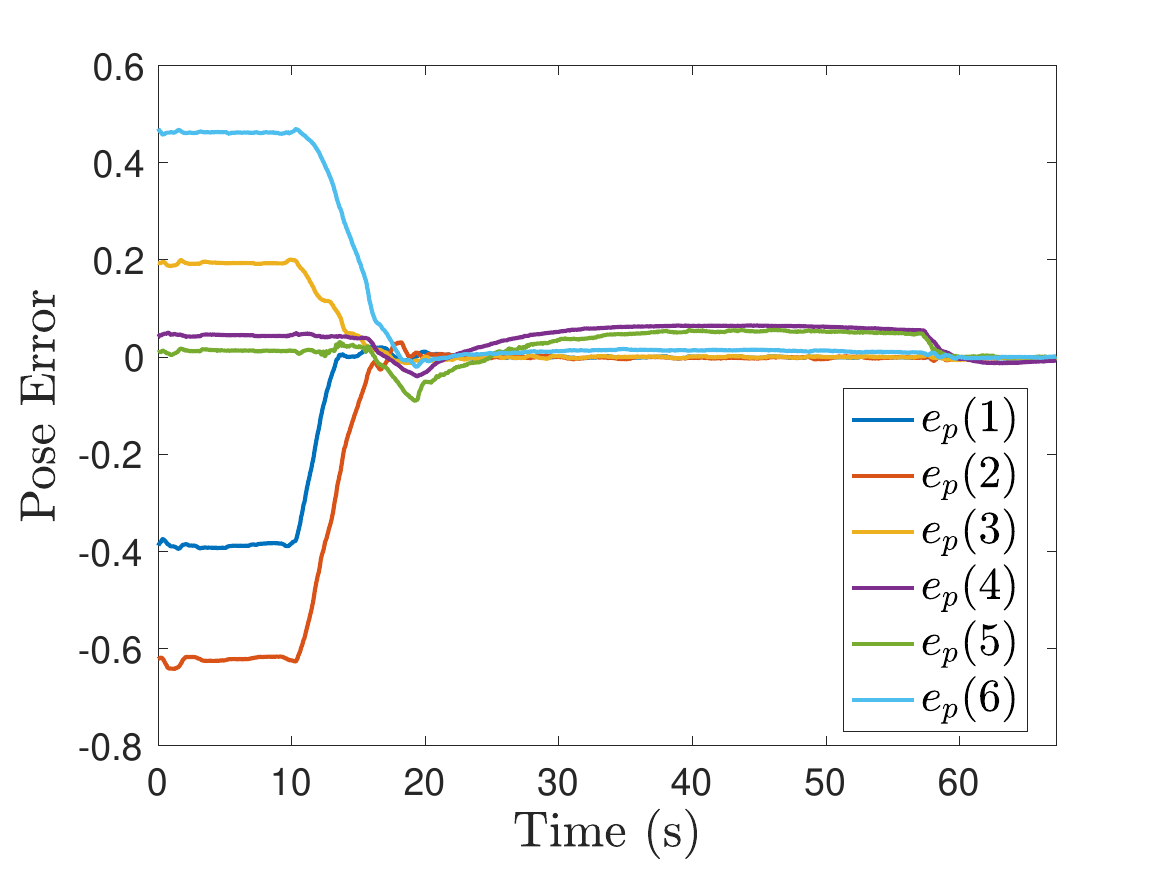}}\hspace{-0.75em}\subfigure[]{\includegraphics[width=0.22\paperwidth]{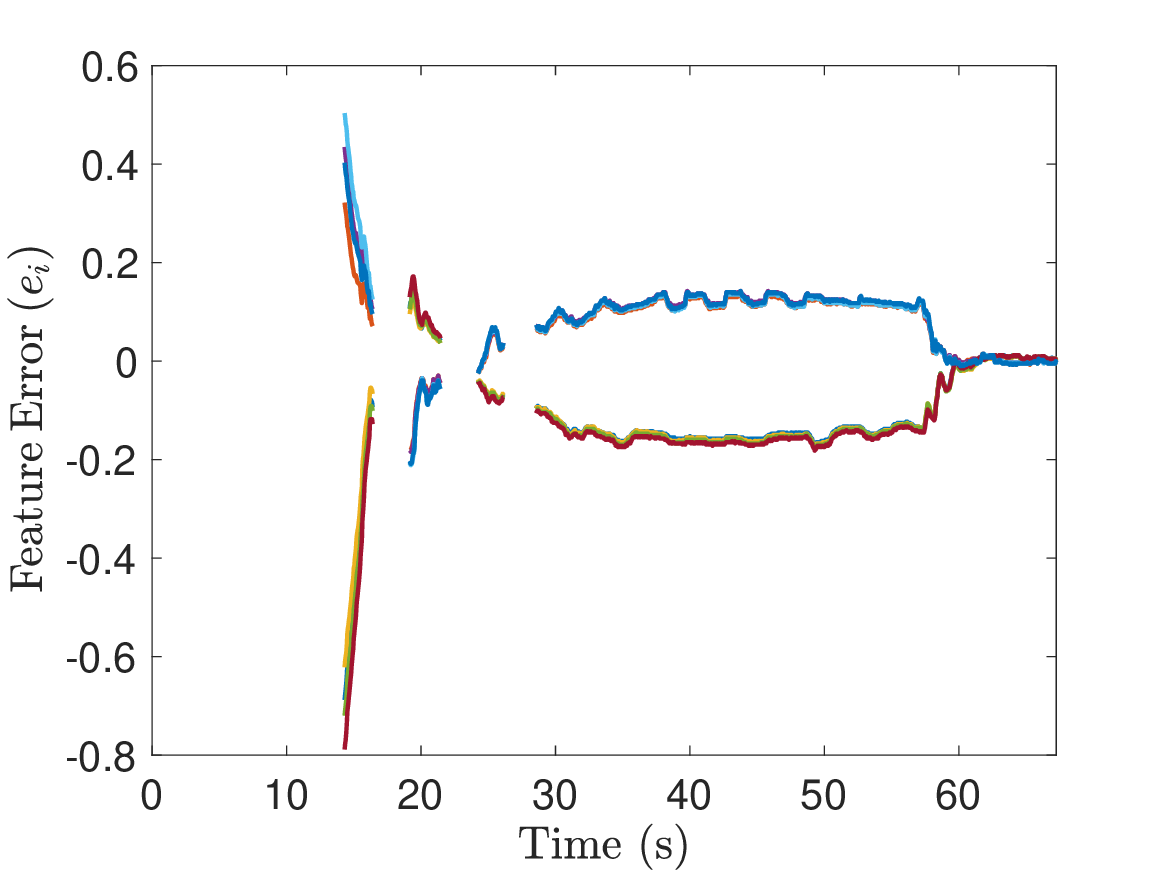}}\hspace{-0.75em}\subfigure[]{\includegraphics[width=0.22\paperwidth]{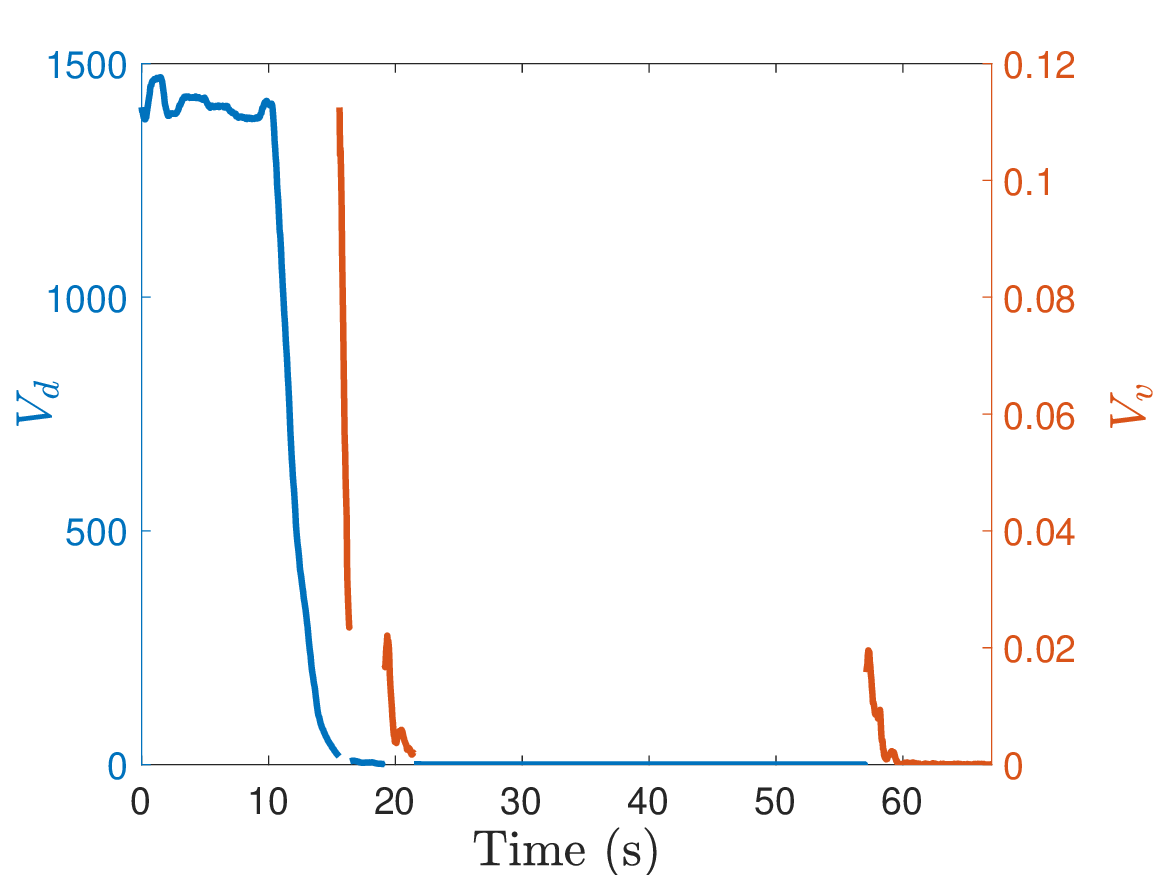}}}\\\vspace{-1.5ex}
\par\end{centering}
\centering{}\caption{Experimental results for the DMP and IBVS with frequent feature occlusion (a) Camera acceleration generated by DMP controller and the IBVS controller in the presence of occlusion with grey dashed-dotted lines showing the switching instances. (b) Pose error $e_{p}(t)$ converges to a bound in the presence of multiple switching instances. (c) Image feature errors $e_{i}(t)$ computed when all the features are visible and satisfy the threshold condition in Algorithm 1. (d) Value of Lyapunov function $V_{\sigma(x,t)}(x(t),t)$ during multiple switching instances, with left y-axis showing the scale for $V_{d}(x(t))$ and right y-axis showing the scale for $V_{v}(x(t),t)$.\label{Experiment2}}
\end{figure*}

In the second experiment, the stability and convergence of DMP-IBVS switched system is tested when there are multiple switches amongst these individual systems. Such a case would occur practically when the feature points are occluded or the features go out of FOV during the IBVS system operation. The feature occlusion is simulated using a piece of paper placed on the ArUco markers covering the camera view. The parameters of the DMP-IBVS switching algorithm are selected empirically as follows: $\mu=10.67$, the decay rates are $\underline{\beta}=0.77$ and $\epsilon = 0.01\underline{\beta}$. The average dwell time is calculated as $\tau_{a}=13.82\:\mathrm{s}$. The individual DMP and IBVS parameters are the same as those from Experiment 1. The constants are selected as $\bar{N} = 5$ and $N_{0}=2$, which allows the total time between $\bar{N}$ switches to be $3\tau_{a}$. This also implies that if the first 4 switches between IBVS and DMP controllers are fast, then DMP-IBVS system ensures that the average dwell time condition is met by keeping the DMP controller active for longer time before the $5^{\mathrm{th}}$ switch occurs. The results of Experiment 2 are summarized in Figs. \ref{Experiment2}(a)-(d). Fig. \ref{Experiment2}(a) shows the acceleration generated by the switched system with the switching instants marked with grey dashed-dotted lines. The switching occurs at the following time instances $t=15.56\:\mathrm{s}$, $t=16.44\:\mathrm{s}$, $t=19.12\:\mathrm{s}$, $t=21.48\:\mathrm{s}$ and $t=57.08\:\mathrm{s}$ with the DMP controller active between $0-15.56\:\mathrm{s}$. 
Fig. \ref{Experiment2}(b) shows the convergence of the pose error to a bound for the switched system using Algorithm 1. As verified by Theorem 3, the pose error is continuous despite the switching between the DMP and IBVS subsystems. Fig. \ref{Experiment2}(c) shows the exponential decay of the image feature errors when the IBVS controller is active and the conditions of Algorithm 1 are met. It can be seen that, although all the features are visible and the feature error is below the selected threshold, the DMP controller is active between $t=21.48-57.08\mathrm{s}$ to compensate for the average dwell time to ensure the stability despite fast initial switching. Fig. \ref{Experiment2}(d) shows the asymptotic convergence of the Lyapunov function $V_{\sigma(x,t)}(x(t),t)$ for the switched system. The result of Theorem 3 is verified by the stability of the Lyapunov functions as shown in Fig. \ref{Experiment2}(d). The error bounds obtained for the experiment in Table \ref{tab:error_bound} verify the convergence of the error for the switched system.

\begin{table}[]
    \caption{Error Bound Information for Baxter Pose Regulation Experiments.}
   \label{tab:error_bound}
    \centering
    \begin{tabular}{|c|c|c|c|c|}
    \hline 
    & $\Vert x\Vert$ & $\Vert \rho \Vert$ & $\Vert e_{i}\Vert$ & $\Vert e_{p}\Vert$\tabularnewline
    \hline 
    \hline 
    Experiment 1 & 0.0660 & 0.0240 & 0.0239  & 0.0658 \tabularnewline
    \hline 
    Experiment 2 & 0.0089 & 0.0078 & 0.0078 & 0.0082 \tabularnewline
    \hline 
    \end{tabular}
\end{table}

\subsection{Experiment 3}
\begin{figure}
\begin{centering}
\makebox[0.6\linewidth]{%
\subfigure[]{\includegraphics[width=0.45\columnwidth]{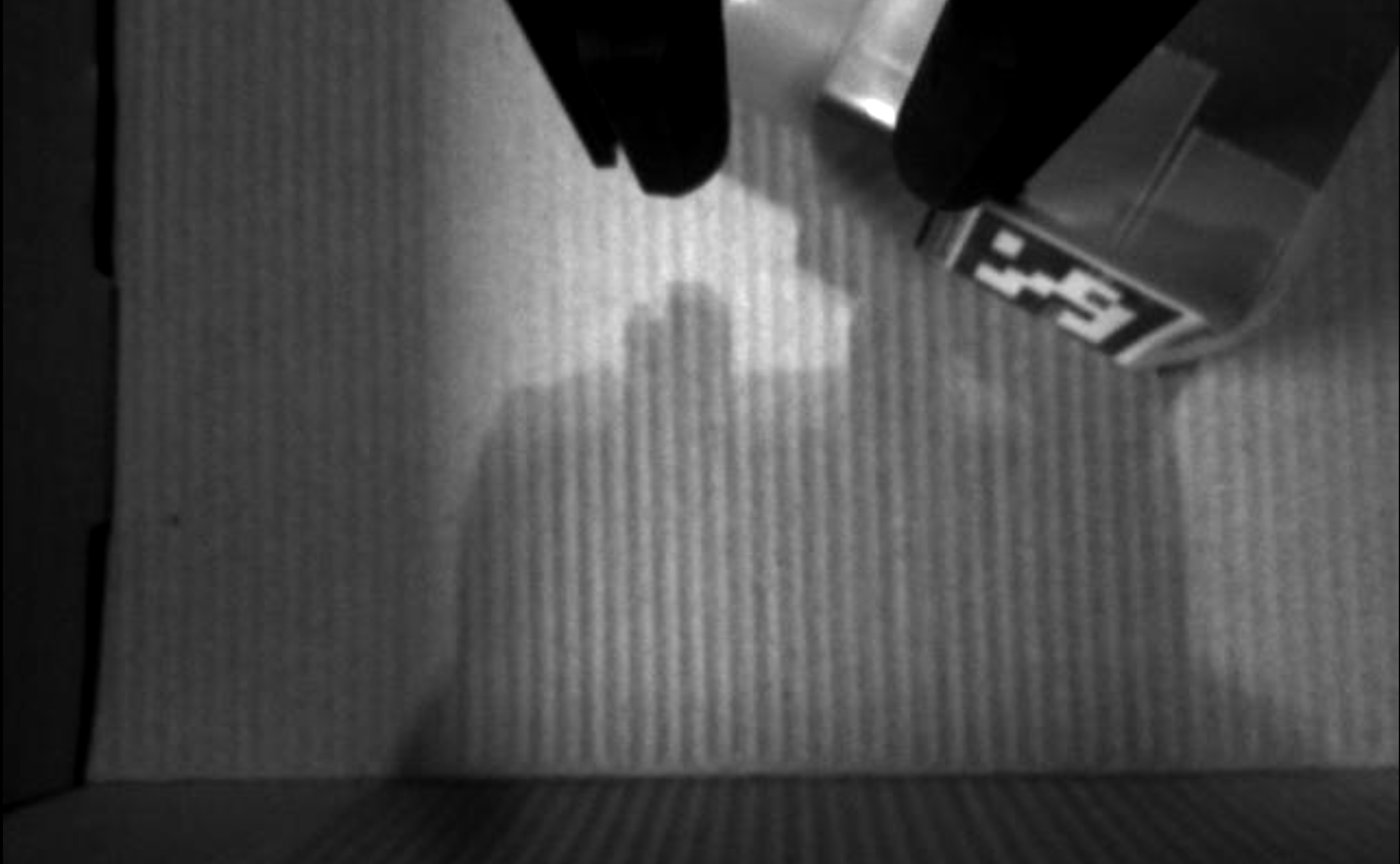}}\hspace{0.4em}
\subfigure[]{\includegraphics[width=0.45\columnwidth]{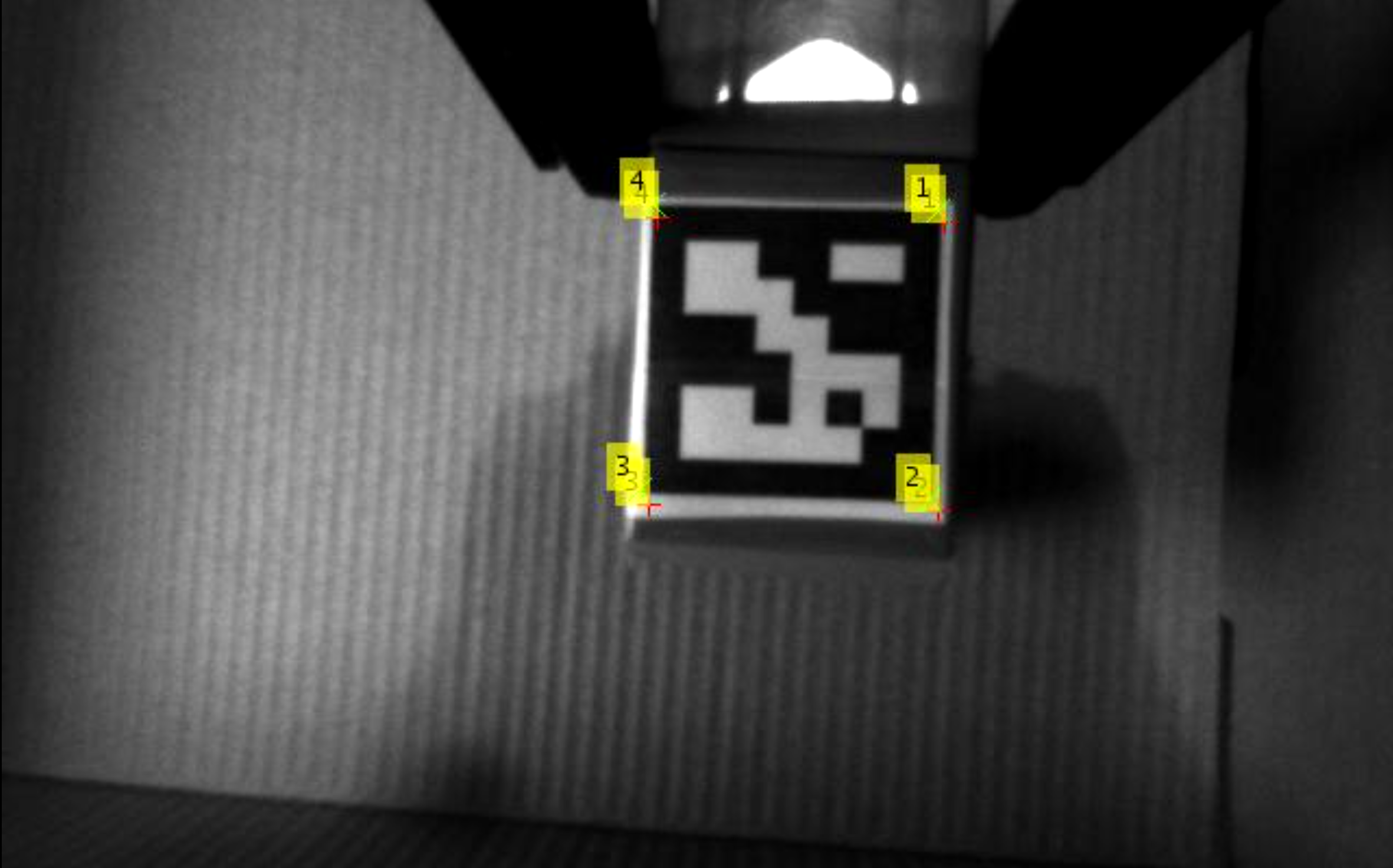}}}
\vspace{-1.5ex}
 
\par\end{centering}
\begin{centering}
\makebox[0.6\linewidth]{%
\subfigure[]{\includegraphics[width=0.5\columnwidth]{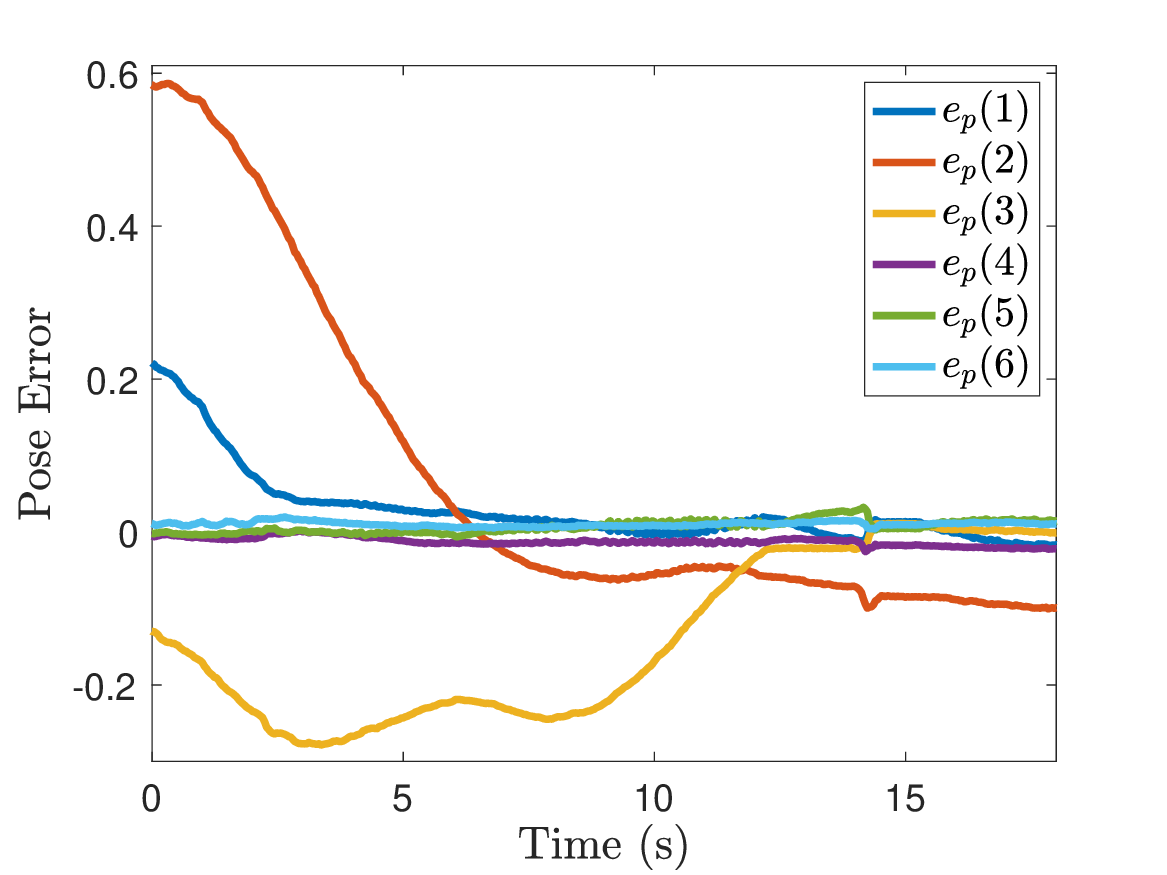}}\hspace{-1em}
\subfigure[]{\includegraphics[width=0.5\columnwidth]{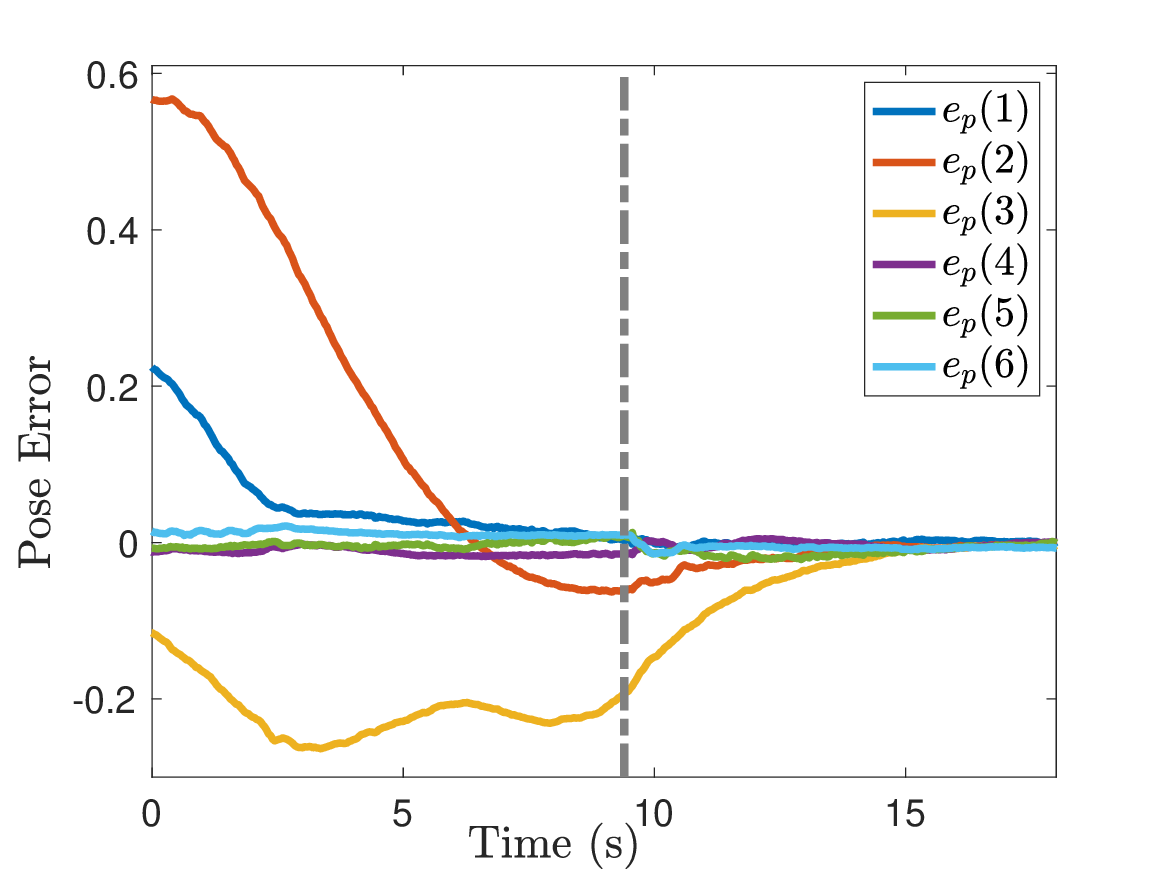}}
}
\par\end{centering}
\centering{}\caption{Experimental results for end-effector positioning to complete the pinch grasping task with changed object location. (a) Failed pinching operation with changed object location using DMP-only system. (b) Successful pinching operation despite changed goal location using the DMP-IBVS switched system. 
(c) Pose error evolution for the DMP. 
(d) Pose error convergence for the DMP-IBVS switched system 
\label{Experiment3}}
\end{figure}

In this experiment, the DMP-IBVS switched system performance is compared with the performance of DMP for a precision pose regulation task of the end-effector when there is an uncertainty in the goal-pose. Precision positioning is required in many fine manipulation tasks such as assembly, small hole insertions, or pinch grasping. A pinch grasp task using parallel jaw grippers is considered in this experiment. For successful pinch grasping the gripper needs to be positioned precisely to a certain pose (i.e., a goal pose) with respect to the object. An ArUco marker is attached to an object which is placed in a box for this task. The DMP-only control and the DMP-IBVS control are implemented. Trajectory data starting from a pose where the object is not visible in the camera FOV due to the presence of the box and ending close to the object before the gripper closes for pinching is collected for training the DMP parameters. The object position in the box is slightly changed during testing before the trajectory is generated using the DMP-only and DMP-IBVS switched system. The changed goal-pose due to the change of object location is unknown to the DMP. Thus, the DMP-only system cannot reach the new pose. In Fig. \ref{Experiment3}(a), the pinch grasp operation fails using the DMP-only system because the gripper could not position itself precisely before the parallel jaw gripper is activated. However, the DMP-IBVS switched system is able to adapt to the change in the goal pose of the gripper using eye-in-hand image feedback once it is available. The successful pinch using the DMP-IBVS switched system is shown in Fig. \ref{Experiment3}(b). Figs. \ref{Experiment3}(c) - (d) show the evolution of the pose error $e_{p}$ computed with respect to the new goal-pose for the DMP-only and DMP-IBVS switched system. As seen by the dashed-dotted grey line in Fig. \ref{Experiment3}(d), the DMP is active until $9.40\:\mathrm{s}$ and then the IBVS subsystem becomes active when image features are visible. The IBVS ensures the convergence of the pose error with respect to the object using image-feedback. The DMP-only system on the other hand is not able to adapt to the change in goal-pose and converges to the goal-pose used during training. As a result, the pose error does not converge to zero as seen from Fig. \ref{Experiment3}(c). The pose error computed in terms of $\Vert e_{p}\Vert$ for DMP system is $0.1105$ whereas for DMP-IBVS switched system is $0.0092$.

\section{Conclusion\label{sec:Conclusion-1}}

In this paper a switching strategy is presented, that utilizes DMP and IBVS methodologies to combine learning-based end-effector control and perception-based control. The Lyapunov stability analysis of the proposed IBVS system yields a UUB result in a local region around the origin, and that of the DMP system yields global asymptotic stability. The switched DMP-IBVS system analysis based on multiple Lyapunov functions shows that the switched system asymptotically converges to a bound. A switching algorithm is presented based on feature visibility while maintaining stability subject to the average dwell time condition. The method is tested on a Baxter research robot using a pose regulation task of the robot's end-effector over an ArUco marker placed on a table. Future work will involve analysis of multiple equilibria when the object is moved and applying the developed method in various manufacturing and space robotics applications.

\bibliographystyle{IEEEtran}
\bibliography{RCL_Complete}

\begin{thebibliography}{10}
\providecommand{\url}[1]{#1}
\csname url@samestyle\endcsname
\providecommand{\newblock}{\relax}
\providecommand{\bibinfo}[2]{#2}
\providecommand{\BIBentrySTDinterwordspacing}{\spaceskip=0pt\relax}
\providecommand{\BIBentryALTinterwordstretchfactor}{4}
\providecommand{\BIBentryALTinterwordspacing}{\spaceskip=\fontdimen2\font plus
\BIBentryALTinterwordstretchfactor\fontdimen3\font minus
  \fontdimen4\font\relax}
\providecommand{\BIBforeignlanguage}[2]{{%
\expandafter\ifx\csname l@#1\endcsname\relax
\typeout{** WARNING: IEEEtran.bst: No hyphenation pattern has been}%
\typeout{** loaded for the language `#1'. Using the pattern for}%
\typeout{** the default language instead.}%
\else
\language=\csname l@#1\endcsname
\fi
#2}}
\providecommand{\BIBdecl}{\relax}
\BIBdecl

\bibitem{Wirepinning}
E.~{Tunstel}, A.~{Dani}, C.~{Martinez}, B.~{Blakeslee}, J.~{Mendoza},
  R.~{Saltus}, D.~{Trombetta}, G.~{Rotithor}, T.~{Fuhlbrigge}, D.~{Lasko}, and
  J.~{Wang}, ``Robotic wire pinning for wire harness assembly automation,''
  \emph{IEEE/ASME International Conference on Advanced Intelligent Mechatronics
  (AIM)}, pp. 1208--1215, 2020.

\bibitem{ijspeert2003learning}
A.~J. Ijspeert, J.~Nakanishi, and S.~Schaal, ``Learning attractor landscapes
  for learning motor primitives,'' in \emph{Advances in Neural Information
  Processing Systems}, 2003, pp. 1547--1554.

\bibitem{chaumette2006visual}
F.~Chaumette and S.~Hutchinson, ``Visual servo control. {I}. basic
  approaches,'' \emph{IEEE Robotics \& Automation Magazine}, vol.~13, no.~4,
  pp. 82--90, 2006.

\bibitem{lopez2009homography}
G.~L{\'o}pez-Nicol{\'a}s, N.~R. Gans, S.~Bhattacharya, C.~Sag{\"u}{\'e}s, J.~J.
  Guerrero, and S.~Hutchinson, ``Homography-based control scheme for mobile
  robots with nonholonomic and field-of-view constraints,'' \emph{IEEE
  Transactions on Systems, Man, and Cybernetics, Part B (Cybernetics)},
  vol.~40, no.~4, pp. 1115--1127, 2009.

\bibitem{gans2011keeping}
N.~R. Gans, G.~Hu, K.~Nagarajan, and W.~E. Dixon, ``Keeping multiple moving
  targets in the field-of-view of a mobile camera,'' \emph{IEEE Transactions on
  Robotics}, vol.~27, no.~4, pp. 822--828, 2011.

\bibitem{Salehi2021ICRA}
I.~Salehi, G.~Rotithor, R.~Saltus, and A.~P. Dani, ``Constrained image-based
  visual servoing using barrier functions,'' in \emph{IEEE International
  Conference on Robotics and Automation}, 2021, pp. 14\,254--14\,260.

\bibitem{Wang2021SMC}
R.~Wang, X.~Zhang, Y.~Fang, and B.~Li, ``Virtual-goal-guided rrt for visual
  servoing of mobile robots with fov constraint,'' \emph{IEEE Transactions on
  Systems, Man, and Cybernetics: Systems}, vol.~52, no.~4, pp. 2073--2083,
  2022.

\bibitem{Huang2021SMC}
Y.~Huang, M.~Zhu, Z.~Zheng, and K.~H. Low, ``Linear velocity-free visual
  servoing control for unmanned helicopter landing on a ship with visibility
  constraint,'' \emph{IEEE Transactions on Systems, Man, and Cybernetics:
  Systems}, vol.~52, no.~5, pp. 2979--2993, 2022.

\bibitem{Ke2017SMC}
F.~{Ke}, Z.~{Li}, H.~{Xiao}, and X.~{Zhang}, ``Visual servoing of constrained
  mobile robots based on model predictive control,'' \emph{IEEE Transactions on
  Systems, Man, and Cybernetics: Systems}, vol.~47, no.~7, pp. 1428--1438,
  2017.

\bibitem{liberzon2003switching}
D.~Liberzon, \emph{Switching in systems and control}.\hskip 1em plus 0.5em
  minus 0.4em\relax Springer Science \& Business Media, 2003.

\bibitem{liberzon1999basic}
D.~Liberzon and A.~S. Morse, ``Basic problems in stability and design of
  switched systems,'' \emph{IEEE Control Systems Magazine}, vol.~19, no.~5, pp.
  59--70, 1999.

\bibitem{hespanha1999stability}
J.~P. Hespanha and A.~S. Morse, ``Stability of switched systems with average
  dwell-time,'' in \emph{Proceedings of the 38th IEEE Conference on Decision
  and Control}, vol.~3, 1999, pp. 2655--2660.

\bibitem{daafouz2002stability}
J.~Daafouz, P.~Riedinger, and C.~Iung, ``Stability analysis and control
  synthesis for switched systems: a switched {L}yapunov function approach,''
  \emph{IEEE Transactions on Automatic Control}, vol.~47, no.~11, pp.
  1883--1887, 2002.

\bibitem{hespanha2004uniform}
J.~P. Hespanha, ``Uniform stability of switched linear systems: Extensions of
  {L}a{S}alle's invariance principle,'' \emph{IEEE Transactions on Automatic
  Control}, vol.~49, no.~4, pp. 470--482, 2004.

\bibitem{lin2009stability}
H.~Lin and P.~J. Antsaklis, ``Stability and stabilizability of switched linear
  systems: a survey of recent results,'' \emph{IEEE Transactions on Automatic
  Control}, vol.~54, no.~2, pp. 308--322, 2009.

\bibitem{branicky1998multiple}
M.~S. Branicky, ``Multiple {L}yapunov functions and other analysis tools for
  switched and hybrid systems,'' \emph{IEEE Transactions on Automatic Control},
  vol.~43, no.~4, pp. 475--482, 1998.

\bibitem{kamalapurkar2018invariance}
R.~Kamalapurkar, J.~A. Rosenfeld, A.~Parikh, A.~R. Teel, and W.~E. Dixon,
  ``Invariance-like results for nonautonomous switched systems,'' \emph{IEEE
  Transactions on Automatic Control}, vol.~64, no.~2, pp. 614--627, 2018.

\bibitem{Astolfi2020_tac}
D.~{Astolfi}, R.~{Postoyan}, and D.~{Nesic}, ``Uniting observers,'' \emph{IEEE
  Transactions on Automatic Control}, vol.~65, no.~7, pp. 2867--2882, 2020.

\bibitem{Prieur2011_tac}
C.~{Prieur} and A.~R. {Teel}, ``Uniting local and global output feedback
  controllers,'' \emph{IEEE Transactions on Automatic Control}, vol.~56, no.~7,
  pp. 1636--1649, 2011.

\bibitem{gans2007stable}
N.~R. Gans and S.~A. Hutchinson, ``Stable visual servoing through hybrid
  switched-system control,'' \emph{IEEE Transactions on Robotics}, vol.~23,
  no.~3, pp. 530--540, 2007.

\bibitem{parikh2016switched}
A.~Parikh, T.-H. Cheng, H.-Y. Chen, and W.~E. Dixon, ``A switched systems
  framework for guaranteed convergence of image-based observers with
  intermittent measurements,'' \emph{IEEE Transactions on Robotics}, vol.~33,
  no.~2, pp. 266--280, 2016.

\bibitem{parikh2018target}
A.~Parikh, R.~Kamalapurkar, and W.~E. Dixon, ``Target tracking in the presence
  of intermittent measurements via motion model learning,'' \emph{IEEE
  Transactions on Robotics}, vol.~34, no.~3, pp. 805--819, 2018.

\bibitem{parikh2017switched}
A.~Parikh, T.-H. Cheng, R.~Licitra, and W.~E. Dixon, ``A switched systems
  approach to image-based localization of targets that temporarily leave the
  camera field of view,'' \emph{IEEE Transactions on Control Systems
  Technology}, vol.~26, no.~6, pp. 2149--2156, 2017.

\bibitem{chen2019switched}
H.~{Chen}, Z.~I. {Bell}, P.~{Deptula}, and W.~E. {Dixon}, ``A switched systems
  approach to path following with intermittent state feedback,'' \emph{IEEE
  Transactions on Robotics}, vol.~35, no.~3, pp. 725--733, 2019.

\bibitem{harris2020target}
C.~G. Harris, Z.~I. Bell, E.~A. Doucette, J.~W. Curtis, and W.~E. Dixon,
  ``Target tracking in the presence of intermittent measurements by a sparsely
  distributed network of stationary cameras,'' in \emph{American Control
  Conference}, 2020, pp. 3491--3496.

\bibitem{zegers2019switched}
H.-Y. Chen, Z.~I. Bell, P.~Deptula, and W.~E. Dixon, ``A switched systems
  framework for path following with intermittent state feedback,'' \emph{IEEE
  Control Systems Letters}, vol.~2, no.~4, pp. 749--754, 2018.

\bibitem{veer2019switched}
S.~{Veer} and I.~{Poulakakis}, ``Switched systems with multiple equilibria
  under disturbances: Boundedness and practical stability,'' \emph{IEEE
  Transactions on Automatic Control}, vol.~65, no.~6, pp. 2371--2386, 2020.

\bibitem{rotithor2020combining}
G.~Rotithor and A.~P. Dani, ``Combining motion primitives and image-based
  visual servo control,'' in \emph{International Symposium on Flexible
  Automation}.\hskip 1em plus 0.5em minus 0.4em\relax American Society of
  Mechanical Engineers, 2020.

\bibitem{Khalil2002}
H.~K. Khalil, \emph{Nonlinear Systems}, 3rd~ed.\hskip 1em plus 0.5em minus
  0.4em\relax Prentice Hall, 2002.

\bibitem{rotithorTCST}
G.~Rotithor, D.~Trombetta, R.~Kamalapurkar, and A.~P. Dani, ``Full- and
  reduced-order observers for image-based depth estimation using concurrent
  learning,'' \emph{IEEE Transactions on Control Systems Technology}, vol.~29,
  no.~6, pp. 2647--2653, 2021.

\bibitem{Spong2006}
M.~W. Spong, S.~Hutchinson, and M.~Vidyasagar, \emph{Robot modeling and
  control}.\hskip 1em plus 0.5em minus 0.4em\relax Wiley New York, 2006,
  vol.~3.

\bibitem{chaumette1998potential}
F.~Chaumette, ``Potential problems of stability and convergence in image-based
  and position-based visual servoing,'' in \emph{The Confluence of Vision and
  Control}.\hskip 1em plus 0.5em minus 0.4em\relax Springer, 1998, pp. 66--78.

\bibitem{han2008control}
D.~Han, Q.~Wei, Z.~Li, and W.~Sun, ``Control of oriented mechanical systems: A
  method based on dual quaternion,'' \emph{IFAC Proceedings Volumes}, vol.~41,
  no.~2, pp. 3836--3841, 2008.

\bibitem{ude2014orientation}
A.~Ude, B.~Nemec, T.~Petri{\'c}, and J.~Morimoto, ``Orientation in {C}artesian
  space dynamic movement primitives,'' \emph{IEEE International Conference on
  Robotics and Automation (ICRA)}, pp. 2997--3004, 2014.

\bibitem{vu2007input}
L.~Vu, D.~Chatterjee, and D.~Liberzon, ``Input-to-state stability of switched
  systems and switching adaptive control,'' \emph{Automatica}, vol.~43, no.~4,
  pp. 639--646, 2007.

\bibitem{klotz2018}
J.~R. {Klotz}, A.~{Parikh}, T.~{Cheng}, and W.~E. {Dixon}, ``Decentralized
  synchronization of uncertain nonlinear systems with a reputation algorithm,''
  \emph{IEEE Transactions on Control of Network Systems}, vol.~5, no.~1, pp.
  434--445, 2018.

\bibitem{ye2020switching}
J.~Ye, S.~Roy, M.~Godjevac, and S.~Baldi, ``A switching control perspective on
  the offshore construction scenario of heavy-lift vessels,'' \emph{IEEE
  Transactions on Control Systems Technology}, vol.~29, no.~1, pp. 470--477,
  2021.

\end{thebibliography}




\end{document}